\documentclass[acmtog,authorversion]{acmart}

\acmJournal{TOG}
\acmYear{2019}
\acmVolume{38}
\acmNumber{6}
\acmArticle{176}
\acmMonth{11}
\acmDOI{10.1145/3355089.3356570}

\setcopyright{rightsretained}

\usepackage{amsmath}
\usepackage{enumitem}
\usepackage{amsfonts} 
\usepackage{booktabs} 
\usepackage{nicefrac} 
\usepackage{microtype}      
\usepackage{bm}

\usepackage{subcaption}
\usepackage{siunitx}
\usepackage{multirow}

\newcommand{\etal}{\textit{et al}.}
\newcommand{\ie}{\textit{i}.\textit{e}.}
\newcommand{\eg}{\textit{e}.\textit{g}.}

\definecolor{darkred}{rgb}{0.8,0,0}
\newcommand{\rev}[1]{{#1}}
\newcommand{\cond}[1]{#1}

\DeclareMathOperator*{\argmin}{arg\,min}
\DeclareMathOperator{\E}{\mathbb{E}}

\setcitestyle{square}

\begin{document}

\title{DeepRemaster: Temporal Source-Reference Attention Networks for Comprehensive Video Enhancement}

\author{Satoshi Iizuka}
\affiliation{%
  \institution{University of Tsukuba}
  \city{Tsukuba}
  \country{Japan}}
\email{iizuka@cs.tsukuba.ac.jp}

\author{Edgar Simo-Serra}
\affiliation{%
  \institution{Waseda University / JST PRESTO}
  \city{Shinjuku}
  \country{Japan}}
\email{ess@waseda.jp}

\begin{teaserfigure}
\includegraphics[width=\linewidth]{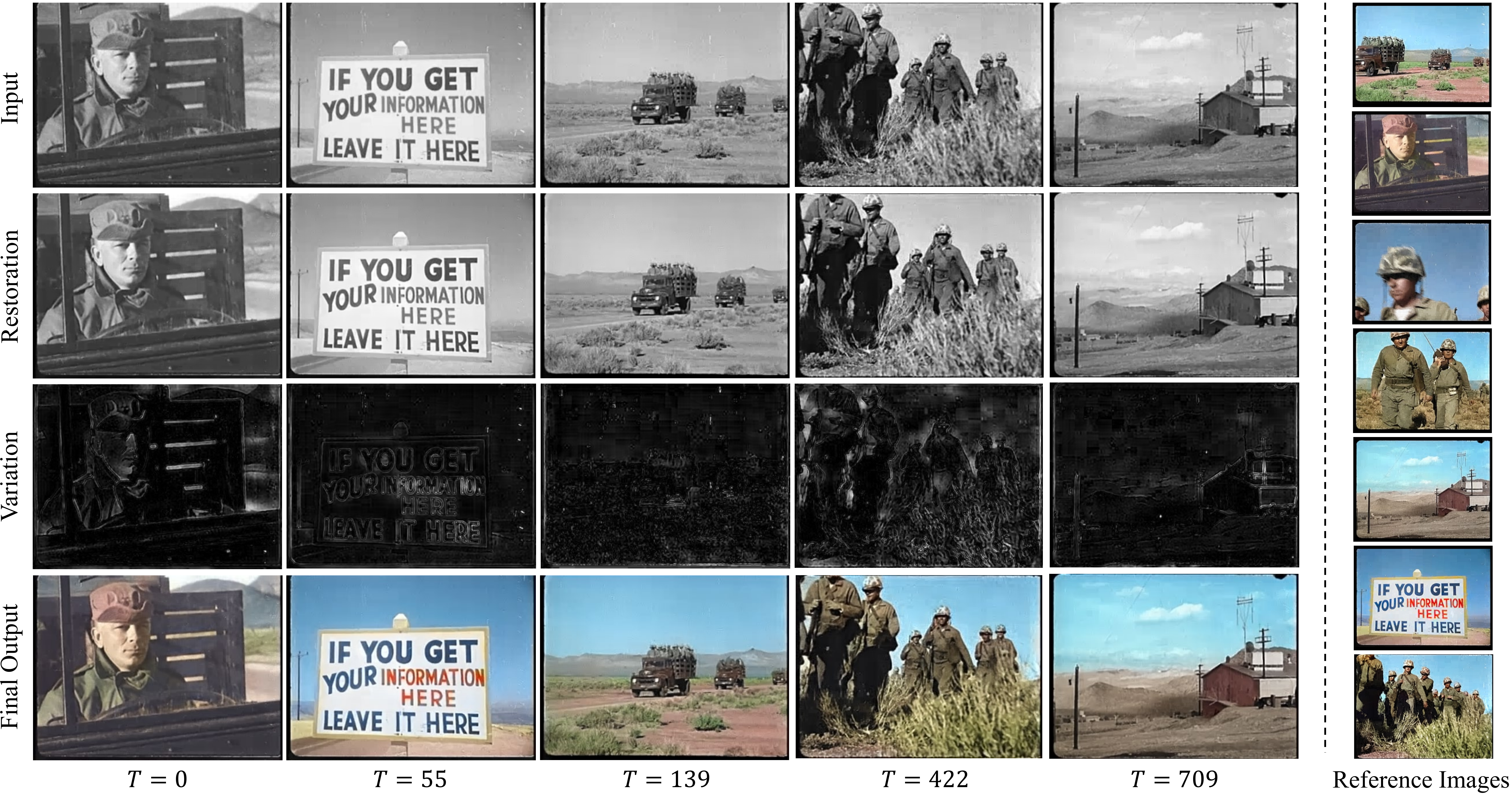}
\vspace{-7mm}
\caption{\textbf{Vintage film remastering results.} Our approach is able to remaster 700
frames of video using only 6 reference color images in a single processing
step. \rev{The first row shows various frames from the input video, the second row shows the
restored black and white frames, the third row shows the variation between the
input and restored black and white frames, and the fourth row shows the final
colorized output.  We show the reference color images used on the right. Using
source-reference attention, our model automatically matches similar regions to
the reference color images, and using self-attention with temporal convolutions
it is able to enforce temporal consistency. Our approach is able to restore the
noisy and blurring input, and, afterwards, with the few manually colored
reference images, we are able to obtain a temporally-consistent natural looking
color video. Images are taken from ``A-Bomb Blast Effects'' (1952) and
licensed under the public domain. Figure best viewed in color.}}
\label{fig:teaser}
\end{teaserfigure}

\begin{abstract}
The remastering of vintage film comprises of a diversity of sub-tasks including
super-resolution, noise removal, and contrast enhancement which aim to restore
the deteriorated film medium to its original state. Additionally, due to the
technical limitations of the time, most vintage film is either recorded in
black and white, or has low quality colors, for which colorization becomes
necessary. In this work, we propose a single framework to tackle the entire
remastering task semi-interactively. Our work is based on temporal
convolutional neural networks with attention mechanisms trained on videos with
data-driven deterioration simulation. Our proposed source-reference attention
allows the model to handle an arbitrary number of reference color images to
colorize long videos without the need for segmentation while maintaining temporal
consistency. Quantitative analysis shows that our framework outperforms
existing approaches, and that, in contrast to existing approaches, the
performance of our framework increases with longer videos and more reference
color images.
\end{abstract}

\ccsdesc[500]{Computing methodologies~Image processing}
\ccsdesc[300]{Computing methodologies~Neural networks}

\keywords{remastering, restoration, colorization, convolutional network, source-reference attention}

\maketitle

\section{Introduction}


\rev{Since the invention of motion pictures in the late 19th century, an incredible
amount of hours of film have been recorded and released. However, in addition
to visual artefacts and the low quality of the film technology at the time,
many of the earlier works of significant historical value have suffered from
degradation or been lost. Restoration of such important films, given their
analogue nature, is complicated, with the initial efforts beginning on
restoring the film at a physical level. Afterwards, the content is transferred
to the digital medium, where it is remastered by removing noise and artefacts
in addition to adding color to the film frames. However, such remastering processes
require a significant amount of both time and money, and is currently
done manually by experts with a single film costing in the order of hundreds
of thousands to millions dollars. Under these circumstances, huge industries
such as publishers, TV, and the print industry, which own an enormous quantity of archived
deteriorated old videos, show a great demand for efficient remastering techniques.
In this work, we propose a semi-automatic approach for remastering old black
and white films that have been converted to digital data.}

Remastering an old film is not as simple as using a noise removal
algorithm followed by colorization approach in a pipeline fashion: the noise and
colorization processes are intertwined and affect each other. Furthermore, most
old films suffer from blurring and low resolution, for which increasing the
sharpness also becomes important. We propose a full pipeline for remastering
black and white motion pictures, made of several trainable components which we
train in a single end-to-end framework. By using a careful data
creation and augmentation scheme, we are able to train the model to remaster
videos by not only removing noise and adding color, but also increasing the
resolution and sharpness, and improving the contrast with temporal consistency.

Our approach is based on fully convolutional networks. In contrast to many
recent works that use recursive models for processing
videos~\cite{LiuECCV2018,VondrickECCV2018}, we use temporal convolutions that
allow for processing video frames by taking account information from multiple
frames of the input video at once. In addition, we propose using an attention
mechanism, which we denote as \emph{source-reference} attention, that allows
using multiple reference frames in an interactive manner. In particular, we use
this \emph{source-reference} attention to provide the model with an arbitrary
number of color mages to be used as references when adding color. The model is
able to not only dynamically choose what reference frames to use when
coloring each output frame, but also choose what regions of the reference
frames to use for each output region in a computationally
efficient manner. We show how this approach can be used to remaster long
sequences composing of multiple different scenes (close-up, panorama, etc.),
using an assortment of reference frames as shown in
Fig.~\ref{fig:teaser}. The number of reference frames used is not fixed and it
is even possible to remaster in a fully automatic way by not providing
reference frames. \rev{Additionally, by manually creating and/or colorizing
reference frames, it is possible for the user to control the colorization
results when remastering, which is necessary for practical applications.}

We perform an in-depth evaluation of our approach both quantitatively and
qualitatively, and find the results of our framework to be favorable in
comparison with existing approaches. Furthermore, the performance of our
approach increases on longer sequences with more reference color images, which
proves to be a challenge for existing approaches.
\rev{Our experiments show that using \emph{source-reference}
attention it is possible to remaster thousands of frames with a small set of
reference images in a efficiently with stable and consistent colors.}

To summarize, our contributions are as follows: (1) the first single framework
for remastering vintage film, (2) source-reference attention that can handle an
arbitrary number of reference images, (3) an \rev{example-based} film degradation
simulation approach for generating training data for film restoration, and (4)
an in-depth evaluation with favorable results with respect to existing
approaches and strong baselines. \cond{Models, code, and additional results are made available at \url{http://iizuka.cs.tsukuba.ac.jp/projects/remastering/}.}


\section{Related Work}

\subsection{Denoising and Restoration}

\newcommand{\bsdfig}[1]{\includegraphics[width=0.48\linewidth]{figs/examples_bsd68/#1}}
\newcommand{\oldfig}[1]{\includegraphics[width=0.33\linewidth]{figs/examples_old/#1}}
\begin{figure*}[ht!]
   \centering
   \begin{subtable}[t]{0.32\linewidth}
      \centering
      \setlength{\tabcolsep}{2pt}
      \begin{tabular}{cc}
         \bsdfig{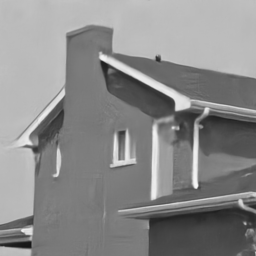} &
         \bsdfig{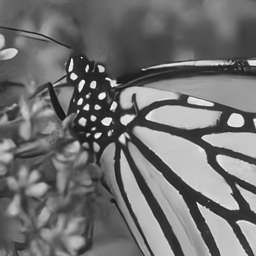} \\
         \bsdfig{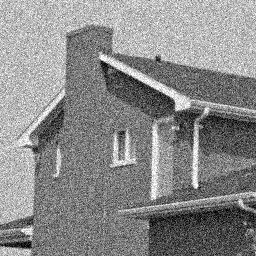} &
         \bsdfig{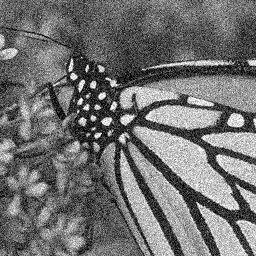} \\
      \end{tabular}
      \vspace{-2mm}
      \caption{Synthetic noisy images.}
   \end{subtable}
   \hspace{5mm}
   \begin{subtable}[t]{0.62\linewidth}
      \centering
      \setlength{\tabcolsep}{2pt}
      \begin{tabular}{ccc}
         \oldfig{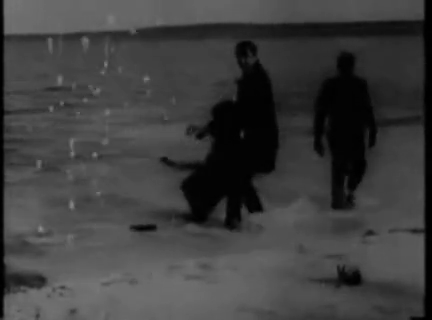} &
            \oldfig{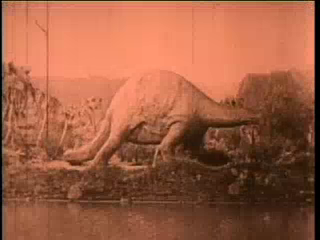} &
            \oldfig{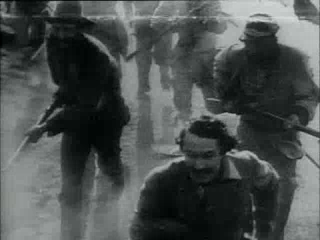} \\
         \oldfig{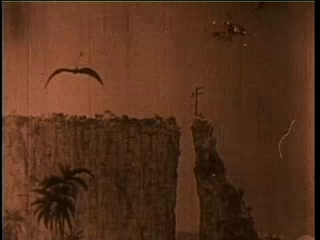} &
            \oldfig{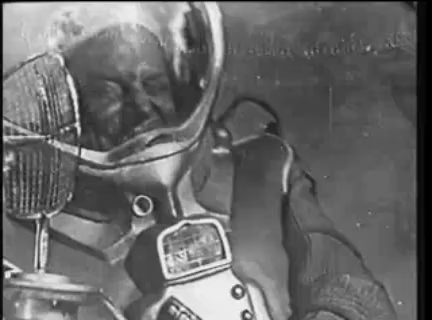} &
            \oldfig{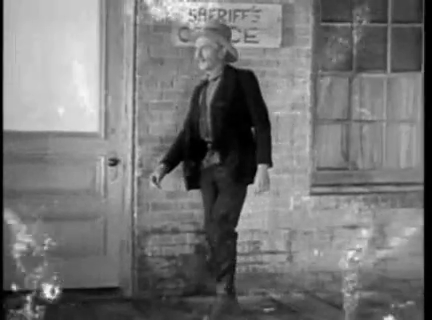} \\
      \end{tabular}
      \setlength{\tabcolsep}{6pt}
      \vspace{-2mm}
      \caption{Vintage black and white movies from the early 20th century.}
   \end{subtable}
   \vspace{-3mm}
   \caption{\textbf{Comparison between denoising and restoration tasks.} (a) Example of
   generated synthetic images for denoising tasks~\cite{MartinICCV2001}. The
   top row shows the original images and the bottom row shows them with added
   Gaussian noise. (b) Example of vintage film which requires restoration.  The
   old movies suffer from a plethora of deterioration issues such as film
   grain noise, scratches, dampness, vignetting, and contrast bleed, which make
   them challenging to restore to their original quality.
   (a) Images are taken from ~\cite{MartinICCV2001}, and (b) videos licensed in the Public Domain.}
   \label{fig:oldfilm}
\end{figure*}

One of the more classical approaches to denoising and restoration is the family
of Block-Matching and 3D filtering (BM3D)
algorithms~\cite{DabovTIP2007,MaggioniTIP2012,MaggioniTIP2014}, which are based on
collaborative filtering in the in the transform domain. Although fairly limited
in the types of noise patterns they can eliminate, these approaches have wide
applicability to both images and video. Besides noise removal, other
restoration related applications such as image super-resolution and
deblurring~\cite{DanielyanTIP2012} have also been explored with the BM3D
algorithm.

More recently, Convolutional Neural Networks have been used for
denoising-type applications, and, in particular, for single
images~\cite{LefkimmiatisCVPR2018,ZhangTIP2018}. However, these generally
assume simple additive Gaussian noise~\cite{LefkimmiatisCVPR2018},
\rev{blurring~\cite{ShiCVPR2016,YuARXIV2018,FanCVPRW2018}},
or JPEG-deblocking~\cite{ZhangTIP2017}, \rev{or are applied to specialized tasks
such as Monte Carlo rendering
denoising~\cite{BakoSIGGRAPH2017,ChaitanyaSIGGRAPH2017,VogelsSIGGRAPH2018}} for
which it is easy to create supervised training data. Extensions for video based
on optical flow and transformer networks have also been
proposed~\cite{KimECCV2018}.  However, restoration of old film requires more
than being able to remove Gaussian noise or blur: it requires being able to
remove film artefacts that can be both local, affecting a small region of the
image, or global, affecting the contrast and brightness of the entire frame, as
shown in Fig.~\ref{fig:oldfilm}. For this it is necessary to create higher
quality and realistic film noise as we propose in our approach.

\subsection{Colorization}

Colorization of black and white images is an ill-posed problem in which there
is no single solution. Most approaches have relied on user inputs, either
in the form of scribbles~\cite{LevinSIGGRAPH2004,HuangACMMM2005}, reference
images similar to the image being
colorized~\cite{ReinhardCGA2001,WelshSIGGRAPH2002,TaiCVPR2005,IronyECRT2005,PitieCVIU2007,WuCGF2013},
or internet queries~\cite{LiuSIGGRAPHASIA2008,ChiaSIGGRAPHASIA2011}.  While
most traditional approaches have focused on solving an optimization problem
using both the input greyscale image and the user provided hints or references
images~\cite{LevinSIGGRAPH2004,AnSIGGRAPH2008,XuSIGGRAPHASIA2013}, recent
approaches have opted to leverage large datasets and employ learning-based
models such as Convolutional Neural Networks (CNN) to colorize images
automatically~\cite{IizukaSIGGRAPH2016,ZhangECCV2016,LarssonECCV2016}.
Analogous to the optimization-based approaches, CNN-based approaches have been
extended to handle user inputs as both
scribbles~\cite{SangkloyCVPR2017,ZhangSIGGRAPH2017}, and a single reference
images~\rev{\cite{HeSIGGRAPH2018,MeyerBMVC2018}}. Our approach, while related to
existing CNN-based methods, extends the colorization to video and an arbitrary
number of reference images, in addition to performing restoration of the video.

\begin{figure}
   \begin{subfigure}{\linewidth}
      \includegraphics[width=\linewidth]{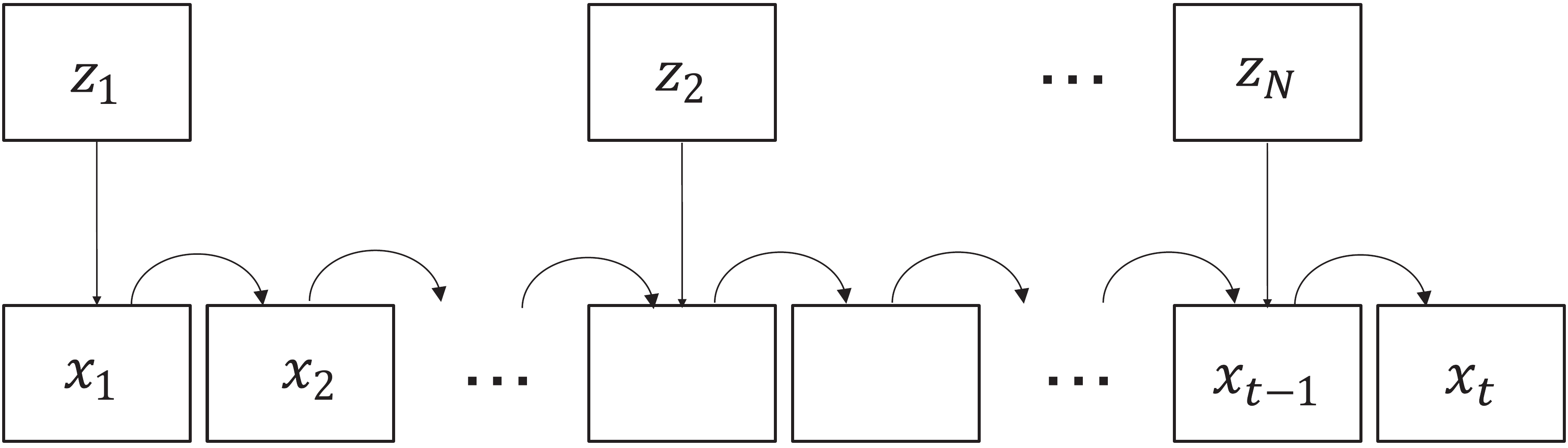}
      \caption{Recursion-based CNN.}
      \vspace{2mm}
   \end{subfigure}
   \begin{subfigure}{\linewidth}
      \includegraphics[width=\linewidth]{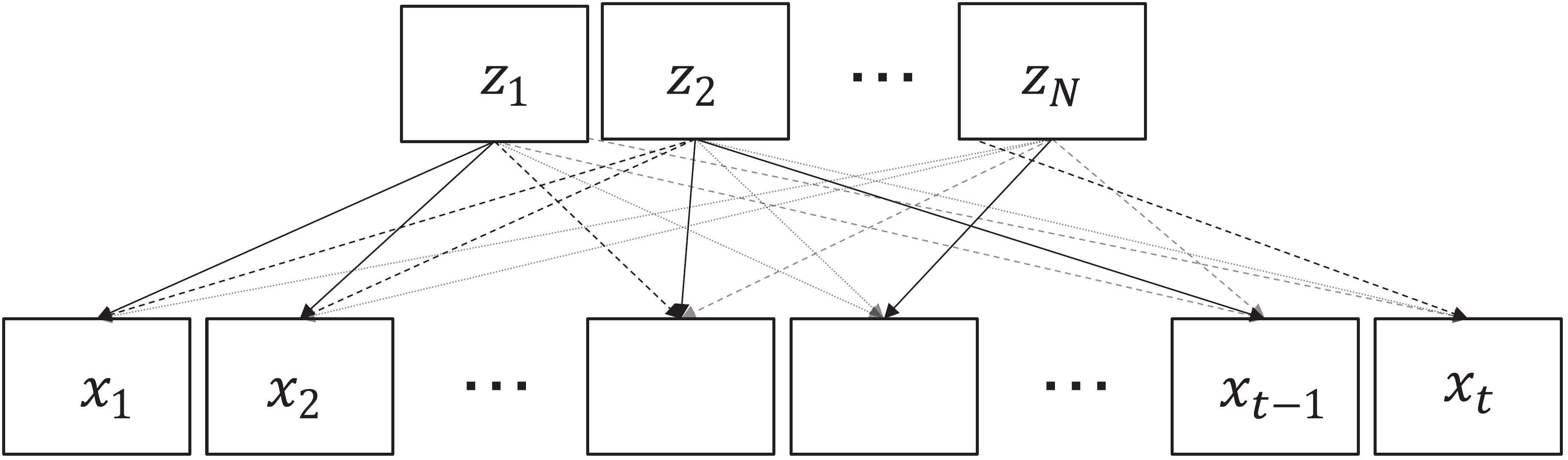}
      \caption{Our Source-Reference Attention-based CNN.}
   \end{subfigure}
   \vspace{-3mm}
   \caption{\textbf{Comparison between recursion-based and attention-based
   Convolutional Neural Networks (CNN) when processing an input video $x$ with
   reference images $z$.} Recursion-based networks simply propagate the
   information frame-by-frame, and because of this can not be processed in
   parallel and are unable to form long-term dependencies. Each time a new
   reference image is used, the propagation is restarted, and temporal
   coherency is lost. Source-reference attention-based networks, such as our
   approach, are able to use all the reference information when processing any
   the frames.}
   \label{fig:recursive}
\end{figure}

Related to the current work are Recursive Neural Network (RNN) approaches for
colorizing videos~\cite{LiuECCV2018,VondrickECCV2018}. They process the video
frame-by-frame by propagating the color from an initial colored key frame to rest of
the scene. While this is a simple way to colorize videos, it can fail to
propagate the color when there are abrupt changes in scene. \rev{In particular,
RNN-based methods have the following limitations:
\begin{enumerate}
\item They require the first frame to be colored and cannot use related frames.
\item They are unable to propagate between scene changes, and thus require precise
scene segmentation. This doesn't allow handling scenes that alternate back and
forth, as commonly done in movies, which end up requiring many additional colorized
references.
\item Once they make an error they continue amplifying it. This severely limits
the number of frames that can be propagated.
\end{enumerate} }
In contrast to RNN-based approaches, our approach is able to handle
multiple scenes or entire videos seamlessly as shown in
Fig.~\ref{fig:recursive}. Instead of using a RNN, we use a CNN with temporal
convolutions and attention, which allows incorporating non-local information
from multiple input frames to colorize a single output frame.

\begin{figure*}[th!]
   \includegraphics[width=\linewidth]{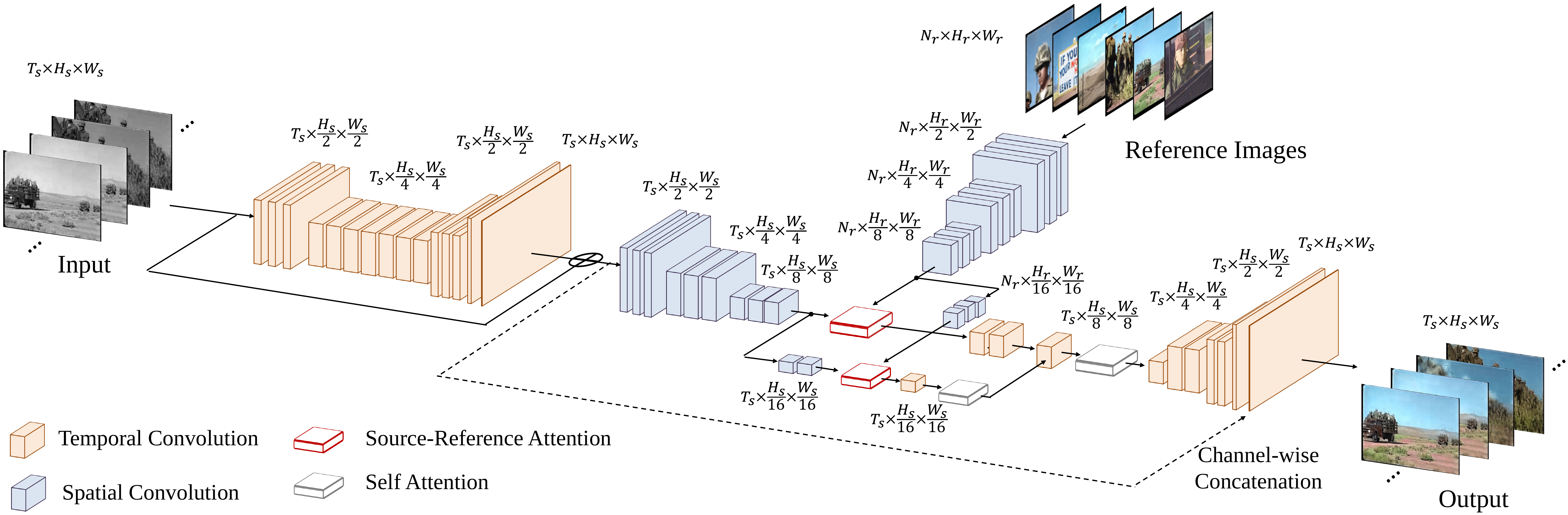}
   \vspace{-4mm}
   \caption{\textbf{Overview of the proposed approach.} The model is input a sequence of
   black and white images which are restored using a pre-processing network and
   used as the luminance channel of the final output video. Afterwards, a
   source-reference network uses an arbitrary number of reference color images
   in conjunction with the output ofthe pre-processing network to produce the
   final chrominance channels of the video. Source-reference attention is used
   to allow the model to employ the color of similar regions in the reference
   color images when colorizing the video. The output of the model is a
   remastered video of the input.}
   \label{fig:overview}
\end{figure*}

\subsection{Attention}

Attention mechanisms for neural networks were original developed for Natural
Language Translation (NLT)~\cite{BahdanauICLR2015}. Similar to human attention,
attention for neural network allows the model to focus on different parts of
the input. For NLT, attention allows to find a mapping
between the input language words and the output language words, which can be in
different orders. For natural language processing, many different variants have
been proposed such as global and local attention~\cite{LuongEMNLP2015},
self-attention~\cite{ChengEMNLP2016,ParikhEMNLP2016} with large-scale
studies being performed~\cite{BritzEMNLP2017,VaswaniNIPS2017}.

Computer vision has also seen applications of attention for caption generation
of images~\cite{XuICML2015}, where the generation of each word in the caption
can focus on different parts of the image using attention. Parmar
\etal~\shortcite{ParmarICML2018} proposed using self-attention for image generation
where pixels locations are explicitly encoded. This was later simplified
in~\cite{ZhangARXIV2018} to not need to explicitly encode the pixel locations.
More related to our approach is the extension of self-attention to videos
classification~\cite{WangCVPR2018}, in which the similarity of objects in the
different video frames is computed with a self-attention mechanism. This is
shown to improve the classification results of videos. Our approach is based on the
same concept, however, we extend it to compute the similarity between the input
video frames and an arbitrary number of reference images.

\section{Approach}

Our approach is based on fully convolutional networks, which are a variant of
convolutional neural networks in which only convolutional layers are employed.
This allows processing images and videos of any resolution.  We employ a mix of
temporal and spatial convolution layers, in addition to attention-based
mechanisms that allow us to use an arbitrary number of reference color images
during the remastering. An overview of the proposed approach can be seen in
Fig.~\ref{fig:overview}.

\subsection{Source-Reference Attention}

\begin{figure}[th]
   \includegraphics[width=\linewidth]{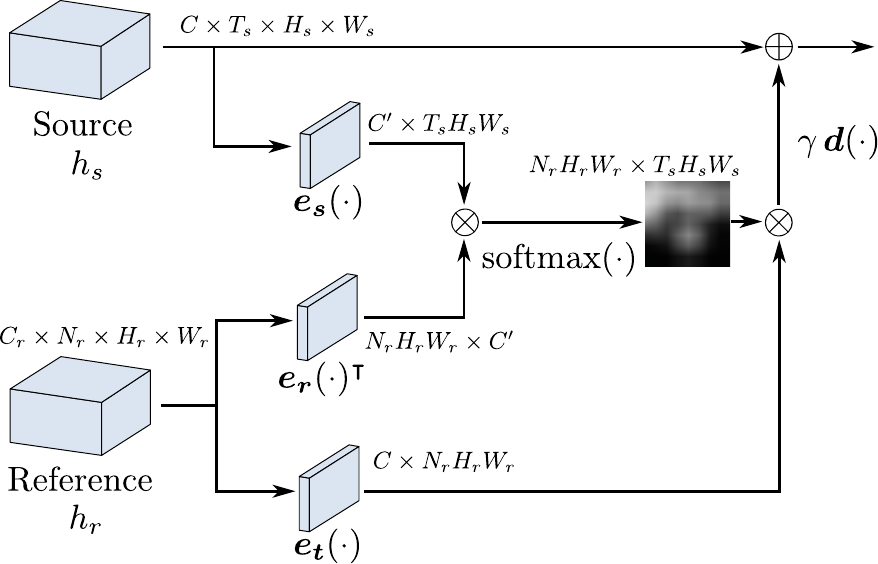}
   \vspace{-4mm}
   \caption{\textbf{Overview of the source-reference attention layer.} This layer takes
   both a set of reference feature maps $h_r$ and a set of source feature maps
   $h_s$ as an input, and outputs a new set of feature maps of the same dimension
   as the source feature maps. This attention allows using non-local features
   from the reference features to perform a transformation of the source
   reference features. This transformation is done while conserving the local
   information similar to a purely convolutional layer. We denote matrix
   multiplication as ``$\otimes$'' and matrix addition as ``$\oplus$''. The
   input and output dimensions of the different components are shown for
   reference.}
   \label{fig:attention}
\end{figure}

We employ source-reference attention to be able to supply an arbitrary number
of reference color images that the model can be use as hints for the
remastering of videos. In particular, source-reference attention layers take as
an input two different variable length volumetric feature maps, one
corresponding to the source data and the other to the reference data, and
allow the model to exploit non-local similarities between the source data and
the reference data. The model can thus use the color from the reference data to
colorize similar areas of the source data.

More formally, let the source data feature representation be $h_s \in
\mathbb{R}^{C\times T_s\times H_s\times W_s}$ with $C$ channels, $T_s$ frames
of height $H_s$ and width $W_s$, and let the reference data features
be $h_r\in\mathbb{R}^{C_r\times N_r\times H_r\times W_r}$ with
$C_r$ channels, $N_r$ maps of height $H_r$ and width $W_r$. The
source-reference attention layer $A_{sr}(\cdot,\cdot)$ can be defined as
\begin{equation}
A_{sr}(h_s, h_r) = h_s + \gamma\,\bm{d}\Big(\bm{e_t}(h_r) \;\text{softmax}\big(\,\bm{e_r}(h_r)^\intercal \; \bm{e_s}(h_s)\,\big)\Big) \quad,
\end{equation}
\noindent where $\gamma\in\mathbb{R}$ is a learnt parameter and
\begin{align}
\bm{e_s}&\colon\mathbb{R}^{C\times T_s\times H_s\times W_s}\to\mathbb{R}^{C'\times T_s H_s W_s} \nonumber \\
\bm{e_r}&\colon\mathbb{R}^{C_r\times N_r\times H_r\times W_r}\to\mathbb{R}^{C'\times N_r H_r W_r} \nonumber \\
\bm{e_t}&\colon\mathbb{R}^{C_r\times N_r\times H_r\times W_r}\to\mathbb{R}^{C\times N_r H_r W_r}
\end{align}
\noindent are encoding functions that map the input source and reference
feature tensors to matrices with a reduced number of channels, and
$\bm{d}\colon\mathbb{R}^{C\times T_s H_s W_s}\to
\mathbb{R}^{C\times T_s\times H_s\times W_s}$ is a decoding function that
simply reshapes the tensor without modifying any values. For the encoding
functions, we use temporal convolution operators with $1\times 1\times 1$-pixel
kernels followed by reshaping to the appropriate output dimensions. A visual
overview of the source-reference attention layer is shown in
Fig.~\ref{fig:attention}.

Note that if reference data features are not provided, the output of the
source-reference attention layer becomes simply the source data features.
Furthermore, in the case the same features are used for both the source and
reference features, the source-reference attention layer becomes a
self-attention layer similar to the implementation of~\cite{ZhangARXIV2018},
except using temporal convolutions instead of spatial convolutions for the
encoders, and the dot-product implementation of~\cite{WangCVPR2018}, where a
single multiplicative parameter $\gamma$ is used in place of a convolution
operator in the decoder. To reduce the computational burden of the attention,
we set $C'=\nicefrac{C}{8}$ unless mentioned otherwise.

\subsection{Model}

The model is composed fundamentally of two trainable parts: a pre-processing
network, and a source-reference network. Both are fully differentiable and
trained together in an end-to-end fashion. We follow the best practices of
fully convolutional networks by having each convolution layer consist of a
convolution operator, followed by a Batch Normalization (BN)
layer~\cite{IoffeICML2015}, and a Exponential Linear Unit (ELU) activation
function~\cite{ClevertICLR2016}, unless specified otherwise. Unless specified
otherwise, all convolutions operate in the temporal domain, with spatial
convolution operators using kernel of size $1\times 3\times 3$, and temporal
convolution operators using kernel of size of $3\times 3\times 3$. All layers
use padding so that the output is the same size as the input .The resolution is
decreased with layers that used strides of $1\times 2\times 2$-pixels, and
increased with trilinear up-sampling before the convolution layers when
necessary.  A full overview of the model can be seen in
Fig.~\ref{fig:overview}.

\subsubsection{Pre-Processing Network}

\begin{table}
   \caption{\textbf{Overview of the pre-processing model architecture.} We abbreviate
   Temporal Convolution with ``TConv.''. Layer irregularities are specified in
   the notes column. When the same layer is repeated several times
   consecutively, we indicate this with the number of times in parenthesis.}
   \label{tbl:model1}
   \vspace{-2mm}
   \centering
   \setlength{\tabcolsep}{4pt}
   \begin{tabular}{lcp{3cm}}
      \toprule
      Layer Type & Output Resolution & Notes \\
      \midrule
      Input & $1 \times T_s \times W_s \times H_s$ & Input greyscale image \\[2mm]

      TConv. & $64 \times T_s \times \nicefrac{W_s}{2} \times \nicefrac{H_s}{2}$ & Replication padding, spatial stride of 2 \\
      TConv. ($\times 2$) & $128 \times T_s \times \nicefrac{W_s}{2} \times \nicefrac{H_s}{2}$ & \\[2mm]

      TConv. & $256 \times T_s \times \nicefrac{W_s}{4} \times \nicefrac{H_s}{4}$ & Spatial stride of 2 \\
      TConv. ($\times 4$)& $256 \times T_s \times \nicefrac{W_s}{4} \times \nicefrac{H_s}{4}$ & \\[2mm]

      TConv. & $128 \times T_s \times \nicefrac{W_s}{2} \times \nicefrac{H_s}{2}$ & Trilinear upsampling \\
      TConv. ($\times 2$) & $64 \times T_s \times \nicefrac{W_s}{2} \times \nicefrac{H_s}{2}$ & \\[2mm]

      TConv. & $16 \times T_s \times W_s \times H_s$ & Trilinear upsampling \\
      TConv. & $1 \times T_s \times W_s \times H_s$ & TanH output, input is added, and finally clamped to $[0,1]$ range\\
      \bottomrule
   \end{tabular}
   \setlength{\tabcolsep}{6pt}
\end{table}

The pre-processing network is formed exclusively by temporal convolution
layers, and uses a skip connection between the input and output. The main
objective of the pre-processing network is to remove artefacts and noise from
the input greyscale video. The network uses an encoder-decoder architecture in
which the resolution is halved twice and restored to the full size at the end
with trilinear upsampling. A full overview of the pre-processing model
architecture is shown in Table~\ref{tbl:model1}. Most of the processing is done
at the low resolution to decrease the computational burden, and the output of
this network is used as the luminance channel of the final output image.

\subsubsection{Source-Reference Network}

\begin{table*}
\caption{\textbf{Overview of the source-reference model architecture.} This model takes as an
input both the output of the pre-processing model and a set of reference
images. Both these inputs are processed by separate encoders (a), then processed in
two different middle branches corresponding to \nicefrac{1}{16} width and
height (b), and \nicefrac{1}{8} width and height (c), before being decoded to
the chrominance channels of the output video with a decoder (d).  We abbreviate
Spatial Convolutions with ``SConv.'', Temporal Convolutions with ``TConv.'',
and Source-Reference Attention with ``SR Attn.''. \rev{For the source and
reference encoders, we refer to the temporal dimension generically as $T$,
where $T=T_r$ for the reference encoder and $T=T_s$ for the source encoder.} We
specify layer irregularities in the notes column. When the same layer is
repeated several times consecutively, we indicate this with the number of times
in parenthesis.}
\label{tbl:model2}
\vspace{-4mm}
\begin{subtable}[t]{0.48\linewidth}
   \vspace{2mm}
   \caption{Source and Reference Encoders.}
   \label{tbl:model_enc}
   \vspace{-1mm}
   \setlength{\tabcolsep}{4pt}
   \begin{tabular}{lcp{3cm}}
      \toprule
      Layer Type & Output Resolution & Notes \\
      \midrule
      Input & $(1\;\text{or}\;3) \times T \times W \times H$ & 3 channels (RGB) for reference, 1 channel (greyscale) for source \\[2mm]

      SConv. & $64 \times T \times \nicefrac{W}{2} \times \nicefrac{H}{2}$ & Spatial stride of 2 \\
      SConv. ($\times 2$) & $128 \times T \times \nicefrac{W}{2} \times \nicefrac{H}{2}$ & \\[2mm]

      SConv. & $256 \times T \times \nicefrac{W}{4} \times \nicefrac{H}{4}$ & Spatial stride of 2 \\
      SConv. ($\times 2$)& $256 \times T \times \nicefrac{W}{4} \times \nicefrac{H}{4}$ & \\[2mm]

      SConv. & $512 \times T \times \nicefrac{W}{8} \times \nicefrac{H}{8}$ & Spatial stride of 2 \\
      SConv. ($\times 2$)& $512 \times T \times \nicefrac{W}{8} \times \nicefrac{H}{8}$ & \\
      \bottomrule
   \end{tabular}
   \setlength{\tabcolsep}{6pt}
\end{subtable}
\begin{subtable}[t]{0.48\linewidth}
   \vspace{2mm}
   \caption{Middle \nicefrac{1}{16} branch.}
   \label{tbl:model_16}
   \vspace{-1mm}
   \setlength{\tabcolsep}{4pt}
   \begin{tabular}{lcp{3cm}}
      \toprule
      Layer Type & Output Resolution & Notes \\
      \midrule
      SConv. & $512 \times T_s \times \nicefrac{W_s}{16} \times \nicefrac{H_s}{16}$ & Input is source encoder output, spatial stride of 2 \\
      SConv. & $512 \times T_s \times \nicefrac{W_s}{16} \times \nicefrac{H_s}{16}$ & Outputs \nicefrac{1}{16} source \\
      \midrule
      SConv. & $512 \times N_r \times \nicefrac{W_r}{16} \times \nicefrac{H_r}{16}$ & Input is reference encoder output, spatial stride of 2 \\
      SConv. ($\times 2$) & $512 \times N_r \times \nicefrac{W_r}{16} \times \nicefrac{H_r}{16}$ & Outputs \nicefrac{1}{16} reference\\
      \midrule
      SR Attn. & $512 \times T_s \times \nicefrac{W_s}{16} \times \nicefrac{H_s}{16}$ & Uses \nicefrac{1}{16} source and reference as inputs \\
      TConv. & $512 \times T_s \times \nicefrac{W_s}{16} \times \nicefrac{H_s}{16}$ & \\
      Self Attn. & $512 \times T_s \times \nicefrac{W_s}{16} \times \nicefrac{H_s}{16}$ & \\
      \bottomrule
   \end{tabular}
   \setlength{\tabcolsep}{6pt}
\end{subtable}
\begin{subtable}[t]{0.48\linewidth}
   \vspace{2mm}
   \caption{Middle \nicefrac{1}{8} branch.}
   \label{tbl:model_8}
   \vspace{-1mm}
   \setlength{\tabcolsep}{4pt}
   \begin{tabular}{lcp{3cm}}
      \toprule
      Layer Type & Output Resolution & Notes \\
      \midrule
      SR Attn. & $512 \times T_s \times \nicefrac{W_s}{8} \times \nicefrac{H_s}{8}$ & Input is source and reference encoder output \\
      TConv. ($\times 2$) & $512 \times T_s \times \nicefrac{W_s}{4} \times \nicefrac{H_s}{4}$ & \\
      TConv. ($\times 2$) & $512 \times T_s \times \nicefrac{W_s}{4} \times \nicefrac{H_s}{4}$ & Output of the \nicefrac{1}{16} branch is concatenated to the input\\
      Self Attn.& $512 \times T_s \times \nicefrac{W_s}{4} \times \nicefrac{H_s}{4}$ & \\
      \bottomrule
   \end{tabular}
   \setlength{\tabcolsep}{6pt}
\end{subtable}
\begin{subtable}[t]{0.48\linewidth}
   \vspace{2mm}
   \caption{Decoder.}
   \label{tbl:model_dec}
   \vspace{-1mm}
   \setlength{\tabcolsep}{4pt}
   \begin{tabular}{lcp{3cm}}
      \toprule
      Layer Type & Output Resolution & Notes \\
      \midrule
      TConv. & $256 \times T_s \times \nicefrac{W_s}{8} \times \nicefrac{H_s}{8}$ &  \\[2mm]

      TConv. & $128 \times T_s \times \nicefrac{W_s}{4} \times \nicefrac{H_s}{4}$ & Trilinear upsampling \\
      TConv. & $64 \times T_s \times \nicefrac{W_s}{4} \times \nicefrac{H_s}{4}$ & \\[2mm]

      TConv. & $32 \times T_s \times \nicefrac{W_s}{2} \times \nicefrac{H_s}{2}$ & Trilinear upsampling \\
      TConv. & $16 \times T_s \times \nicefrac{W_s}{2} \times \nicefrac{H_s}{2}$ & \\[2mm]

      TConv. & $8 \times T_s \times W_s \times H_s$ & Trilinear upsampling \\
      TConv. & $2 \times T_s \times W_s \times H_s$ & Sigmoid output represents chrominance\\
      \bottomrule
   \end{tabular}
   \setlength{\tabcolsep}{6pt}
\end{subtable}
\end{table*}

The source-reference network forms the core of the model and takes as an input
the output of the pre-processing network along with an arbitrary number of
user-provided reference color images.  Two forms of attention are employed to
allow non-local information to be used when computing the output chrominance
maps: source-reference attention allows information from reference color images
to be used, giving the user indirect control of the colorization; and
self-attention allows non-local temporal information to be used, increasing the
temporal consistency of the colorization. For self-attention, we use the
source-reference attention layer implementation and use the same features
for both the source and reference feature maps. \rev{An overview of the
source-reference model architecture is shown in Table~\ref{tbl:model2}.}

As with the pre-process network, the model is based on a encoder-decoder
architecture, where the resolution is reduced to allow for more efficient
computation and lower memory usage, and restored for the final output.
While temporal convolutions allow for better temporal consistency, they also
complicate the learning and increase the computational burden.  Unlike the
pre-processing network, the source-reference network uses a mix of temporal and
spatial convolutions. In particular, the decoder and $\nicefrac{1}{8}$ middle
branch use temporal convolutions while the encoders of both the input video and
reference images use spatial convolutions, and the $\nicefrac{1}{16}$ middle
branch uses a mixture of both, which we found decreases memory usage and
simplifies the training, while not sacrificing any remastering accuracy.
Furthermore, in the case of the reference color images, there is no temporal
coherency to be exploited by using temporal convolutions as the images are not
necessarily related.

First, the input video and reference images are separately reduced to
$\nicefrac{1}{8}$ of the original width and height in three stages by separate
encoders.
The encoded input video and reference video features are then split into two
branches: one processes the video at $\nicefrac{1}{8}$ width and height, and
one decreases the resolution another stage to $\nicefrac{1}{16}$ of the original
width and height to further process the video. Both branches employ
source-reference attention layers, additional temporal convolution layers, and
self-attention layers. In particular, the $\nicefrac{1}{16}$ branch is
processed with a self-attention layer before being upsampled trilinearly and
concatenated to the $\nicefrac{1}{8}$ branch output. The resulting combined
features are processed with self-attention to be more temporally uniform.
Afterwards, a decoder converts the features to chrominance channels in
three stages \rev{using trilinear upsampling}.
Finally the output of the
network is used as the image chrominance with two channels corresponding to
the $ab$ channels of the $Lab$ color-space, while the output of the
pre-processing network is used as the image luminance corresponding to the $L$
channel.

\section{Training}

We train our model using manually curated supervised training data. In order to
improve both the generalization and quality of the results, we perform large
amounts of both synthetic data augmentation and \rev{example-based} film deterioration.

\subsection{Objective Function}

We train the model in a fully supervised fashion with a linear combination of
two $L_1$ losses. In particular, we use a supervised dataset $\mathcal{D}$
consisting of pairs of deteriorated black and white videos $x$ and restored
color videos split into luminance $y_l$ and chrominance $y_{ab}$ using the
$Lab$ color-space, and reference color images $z$, and optimize the
following expression:
\begin{equation}
\argmin_{\theta, \phi} \E_{(x,y_l,y_{ab},z)\in \mathcal{D}} \|P(x;\theta) - y_l\|
   + \beta\,\|S\big( P(x;\theta), z;\phi\big) - y_{ab}\| \;,
\label{eq:optimization}
\end{equation}
\noindent where $P$ is the pre-processing model with weights $\theta$, $S$ is
the source-reference model with weights $\phi$, and $\beta\in\mathbb{R}$ is a
weighting hyper-parameter.

Training is done using batches of videos with 5 sequential frames
each, that are chosen randomly from the training data. For each 5-frame video,
a random number of color references images $z$ is chosen uniformly from the
$[0,6]$ range. If the number of references is not 0, one of the reference
images is chosen to be from within five neighboring frames of the input frames,
while the remaining reference images are randomly sampled from the whole
training data set.

\subsection{Training Data}

We base our dataset on the YouTube-8M dataset~\cite{AbuElHaijaARXIV2016} which
consists of roughly 8 million videos corresponding to about 500 thousand
hours of video data. The dataset is annotated with 4,803 visual entities
which we do not use. We convert the videos to black and white and corrupt
them, simulating old film degradation, to create supervised training data for
our model.

As YouTube-8M dataset was created by mostly automatically, a large amount of
videos depict gameplay, black and white video, static scenes from fixed
cameras, and unnaturally colored scenes such as clubs with live music. We
randomly select videos from the full dataset and manually annotate them as
suitable for training and evaluating a remastering model. In particular, we end
up with \cond{1,569} videos totalling 10,243,010 frames, of which we use \cond{1,219}
(7,993,132 frames) for training our model, 50 (321,306) for validation, and 300
(1,928,572) for testing.

\subsection{Data Augmentation}

\begin{table}
   \caption{\textbf{Overview of the different types of data augmentation used during
   training.} The target refers to which data is being augmented. Values in
   parenthesis indicate that the same transformation is done jointly to both
   variables, instead of independently. Probability indicates how likely that
   particular transformation is likely to occur, and range is how the
   transformation parameters are sampled. We note that in the case of the input
   video $x$ and target video $y$, the same transformation is done to all the
   frames in the video, while in the case of the reference images $z$, the
   transformation is done independently for each image as they are not related
   to each other.}
   \label{tbl:dataaug}
   \vspace{-2mm}
   \setlength{\tabcolsep}{3pt}
   \begin{tabular}{rcccp{3cm}}
      \toprule
      Name & Target & Prob. & Range & Notes \\
      \midrule
      Horiz. Flip& $(x,y),z$ & 50\%  & - & \\[2mm]

      Scaling    & $(x,y)$ & 100\% & $\mathcal{U}(256,400)$ & Size of the smallest edge (px), randomly crops \\
      Rotation   & $(x,y)$ & 100\% & $\mathcal{U}(-5,5)$ & In degrees \\
      Brightness & $(x,y)$ & 20\%  & $\mathcal{U}(0.8, 1.2)$ & \\
      Contrast   & $(x,y)$ & 20\%  & $\mathcal{U}(0.9, 1.0)$ & \\[2mm]

      JPEG       & $x,z$ & 90\% & $\mathcal{U}(15,40)$ & Encoding quality \\
      Noise      & $x,z$ & 10\% & $\mathcal{N}(0, 0.04)$ & Gaussian\\[2mm]

      Blur       & $x$ & 50\% & $\mathcal{U}(2,4)$ & Bicubic down-sampling \\
      Contrast   & $x$ & 33\% & $\mathcal{U}(0.6, 1.0)$ & \\

      Scaling    & $z$ & 100\%& $\mathcal{U}(256,320)$ & Size of the smallest edge (px), randomly crops \\
      Saturation & $z$ & 10\% & $\mathcal{U}(0.3, 1.0)$ & \\
      \bottomrule
   \end{tabular}
   \setlength{\tabcolsep}{6pt}
\end{table}
We perform large amounts of data augmentation to the input video, ground truth
video, and reference images with two objectives: first, we wish to increase the
generalization of the model to different types of video, and secondly, we want
the model to be able to restore different artefacts which can be commonly found
in the input videos, such as blur or low contrast. This data augmentation is
done in parallel with \rev{example-based} deterioration that further degrades the input
greyscale videos.

We use batches of 5-frame videos with their associated
reference images with a resolution of $256\times 256$ pixels. As data
augmentation, we perform a large amount of transformations that affect the
input video $x$ and ground truth video $y=(y_l,y_{ab})$ together, only the
input video $x$, only the reference images $z$, or any combination of the
previous three. An overview of the different transformations we apply is shown
in Table~\ref{tbl:dataaug}, which include changes to brightness, contrast, JPEG
noise, Gaussian noise, blurring, and saturation.

\subsection{\rev{Example-based} Deterioration}

\newcommand{\dfig}[1]{\includegraphics[width=0.48\linewidth]{figs/deterioration/noise#1}}
\begin{figure}
   \centering
   \setlength{\tabcolsep}{2pt}
   \begin{tabular}{cc}
      \dfig{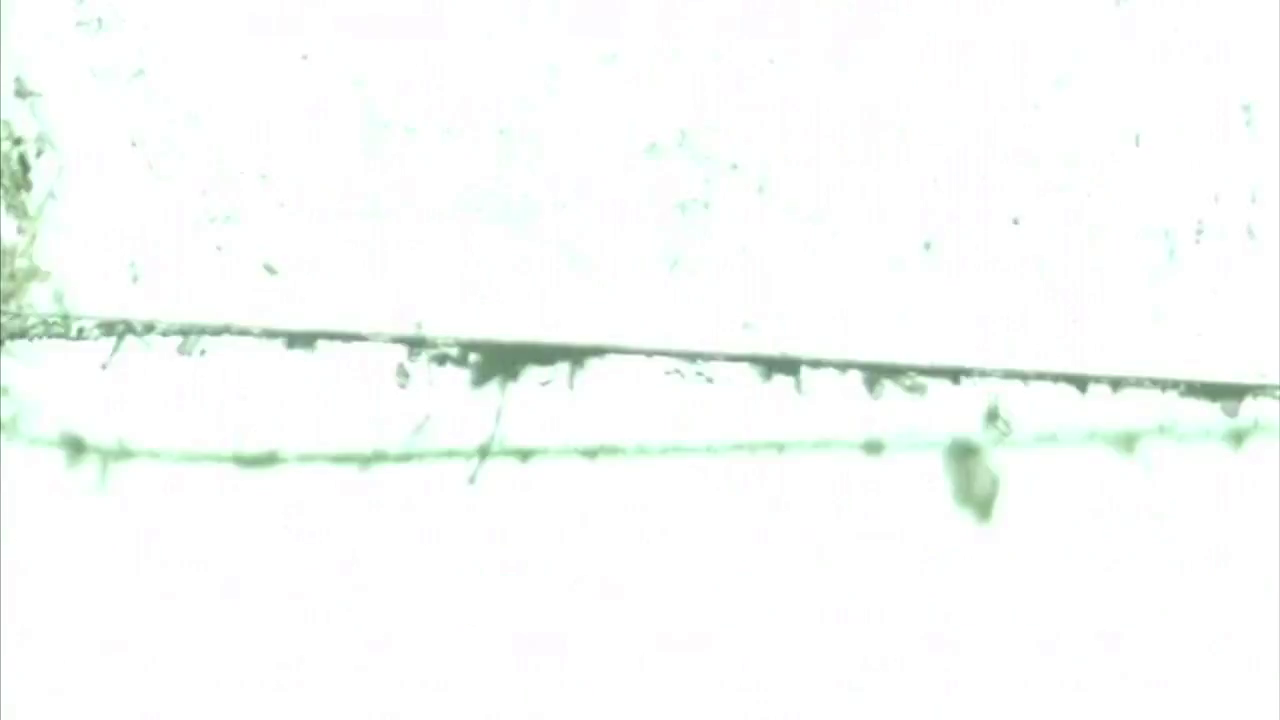} & \dfig{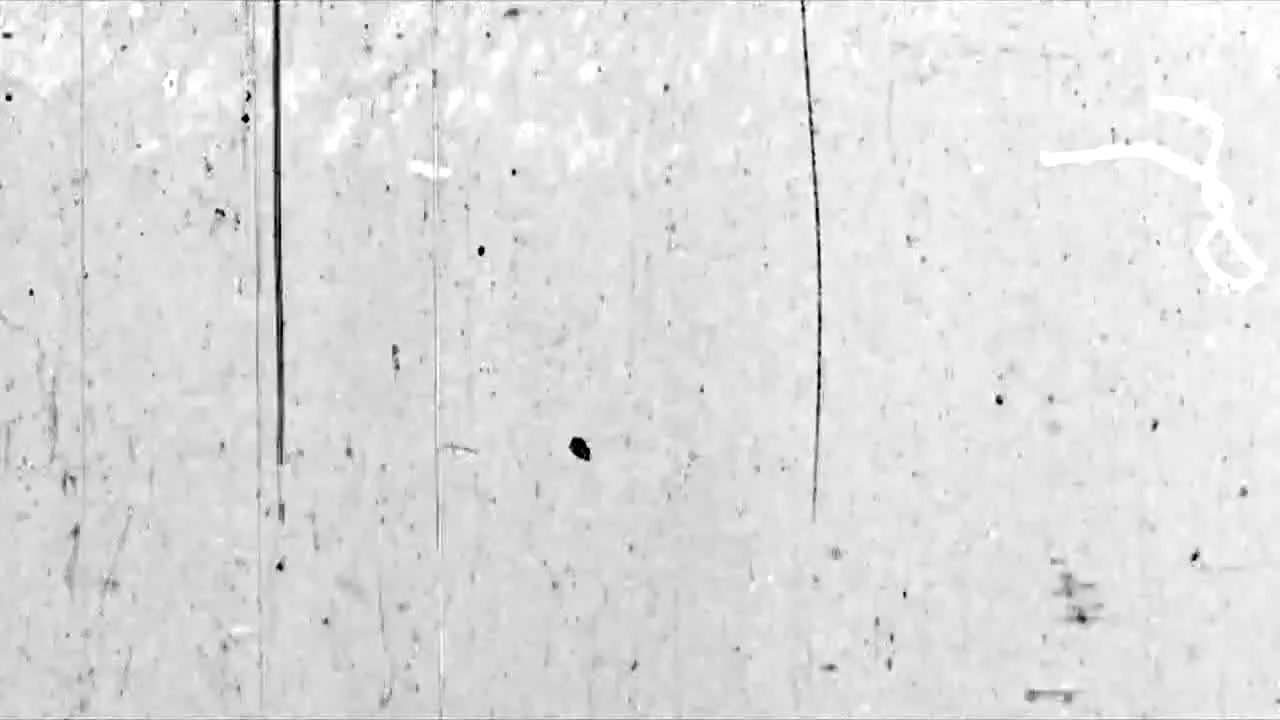} \\
      \dfig{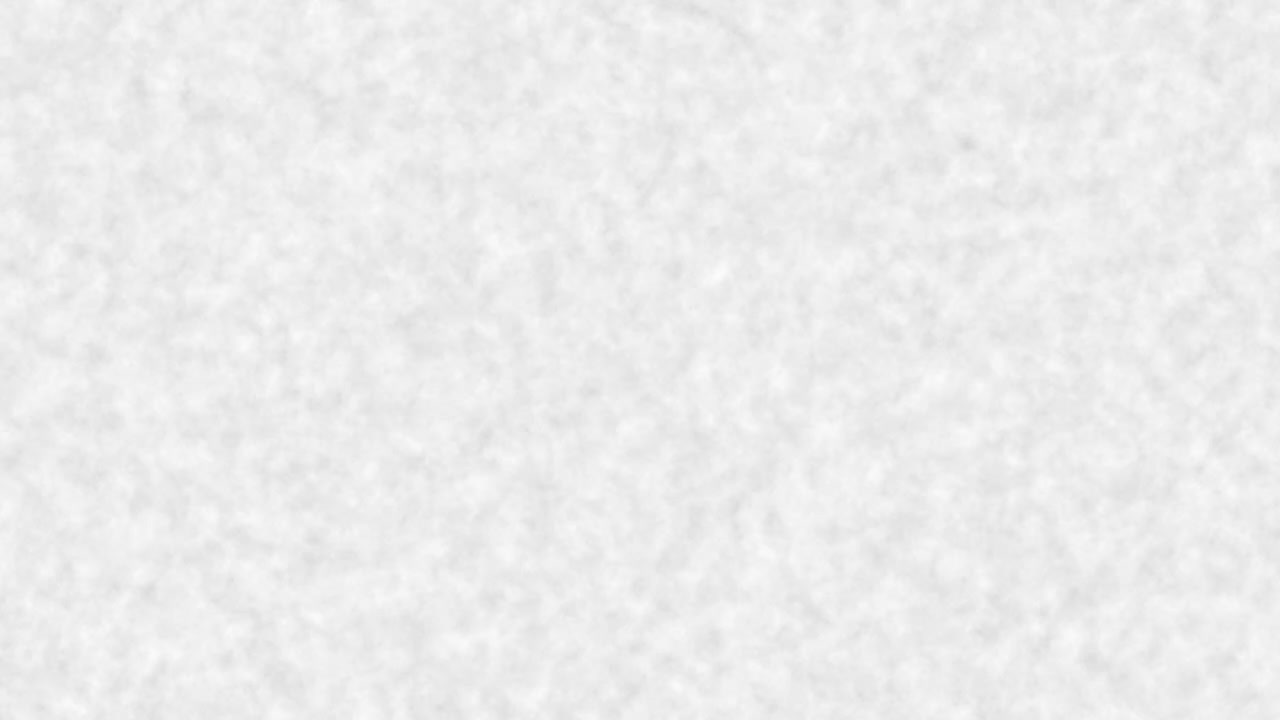} & \dfig{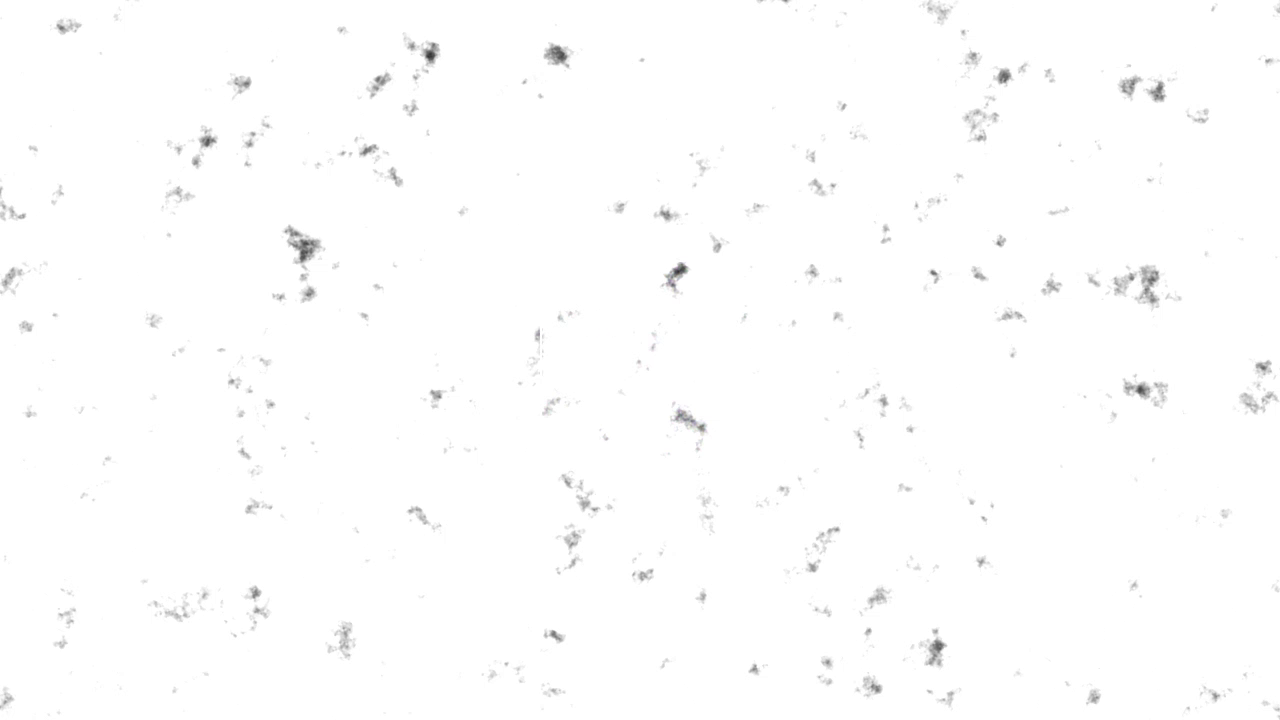} \\
      \dfig{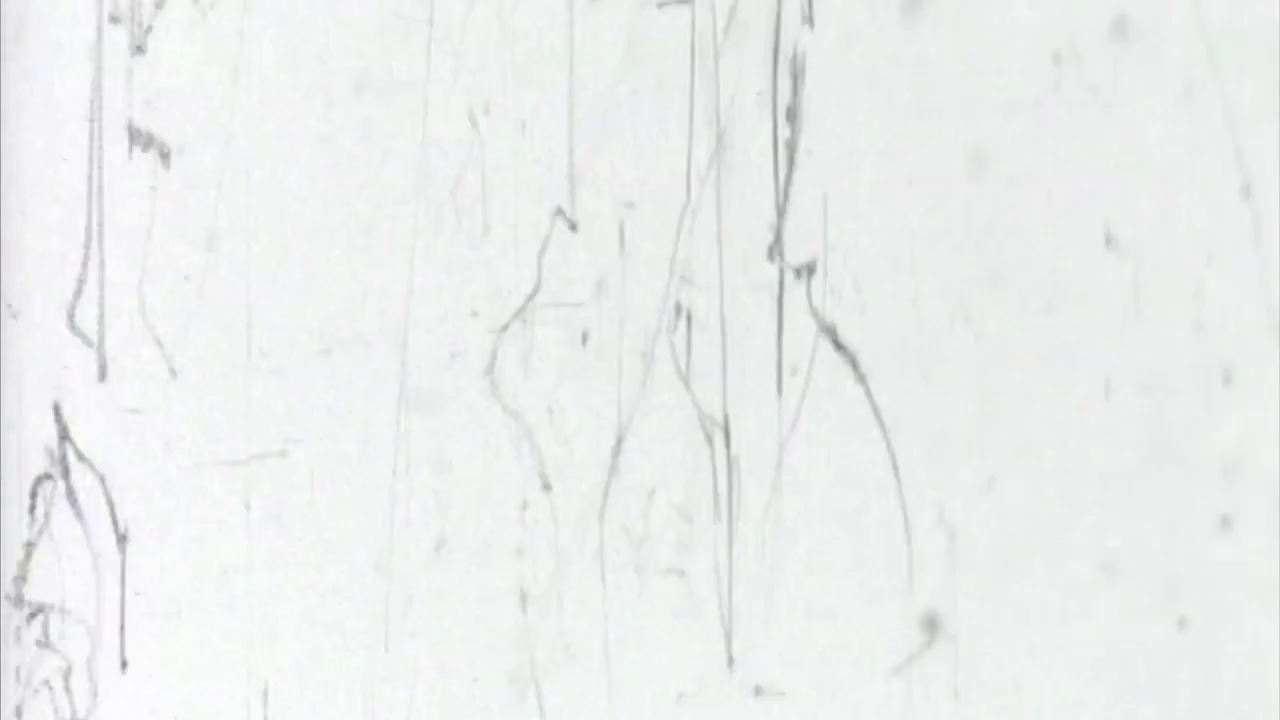} & \dfig{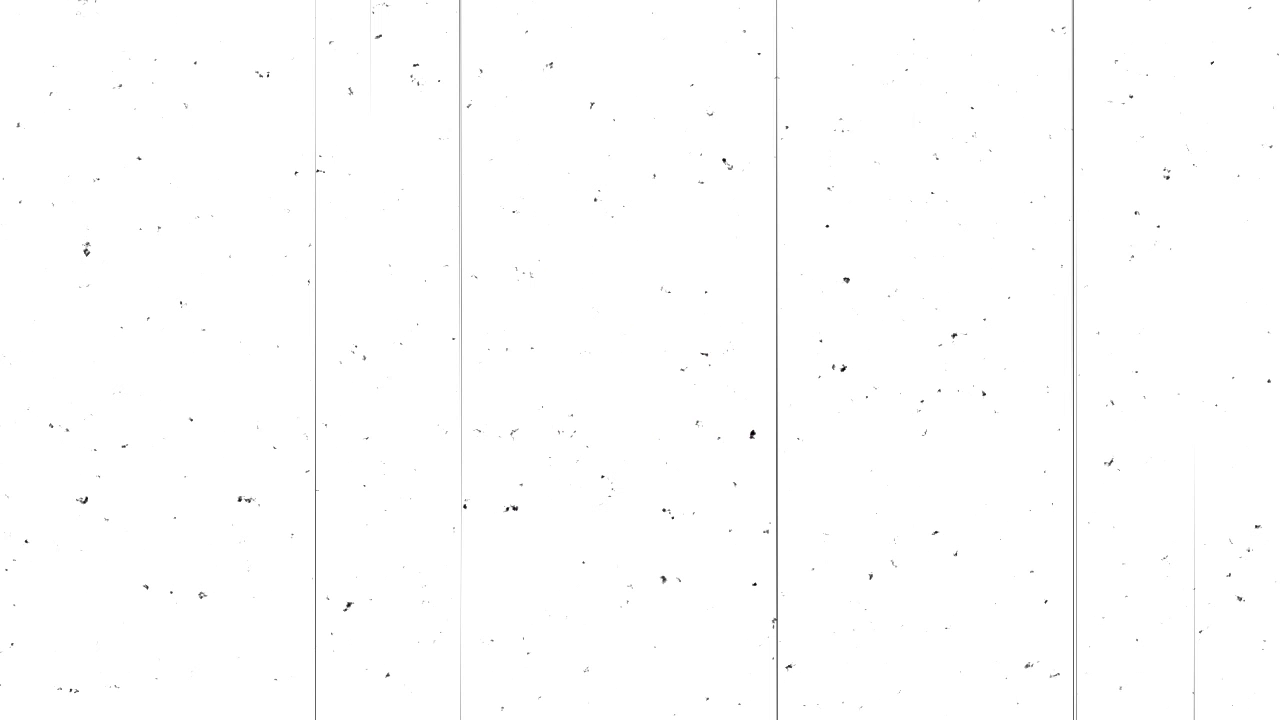} \\
   \end{tabular}
   \setlength{\tabcolsep}{6pt}
   \vspace{-4mm}
   \caption{\cond{\textbf{Example-based deterioration effects.} These effects
   are generated offline and stored as a dataset of images which can then be
   applied to training data inputs as additive noise.}}
   \label{fig:deterioration_examples}
\end{figure}

In addition to all the different data augmentation techniques, we also
simulate deterioration of the film medium from \cond{a dataset of 6,152}
examples, such as fractal noise, grain noise, dust, and scratches. \cond{These
deterioration examples are manually collected by web search using the keywords
``film noise'', and also generated using software such as Adobe After Effects.
For generated noise, fractal noise is used to generate a base
noise pattern, which can then be improved by modifying the contrast,
brightness, and tone curves to obtain scratch and dust-like noise.  In total,
4,340 noise images were downloaded and 1,812 were generated. Some of the
deterioration examples are shown in Fig.~\ref{fig:deterioration_examples}}. In
particular, as these deterioration effects simulate the degradation of the
physical medium which is supporting the film, they are implemented as additive
noise: the noise data is randomly added to the input greyscale video,
independently for each frame.  Furthermore, they are added independently of
each other and combined to create the augmented input videos.

\newcommand{\ddfig}[1]{\includegraphics[width=0.48\linewidth]{figs/deterioration/27387069482#1}}
\begin{figure}
   \centering
   \setlength{\tabcolsep}{2pt}
   \begin{tabular}{cccc}
      \ddfig{}      & \ddfig{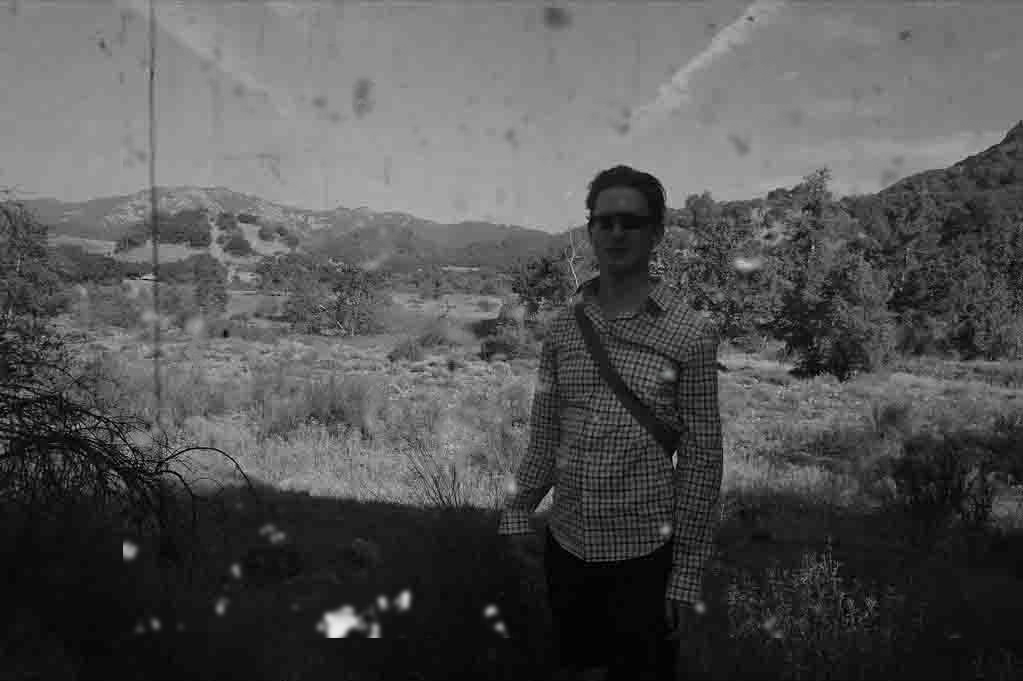} \\
      (a) Original & (b) Deteriorated Example 1 \\[1mm]
      \ddfig{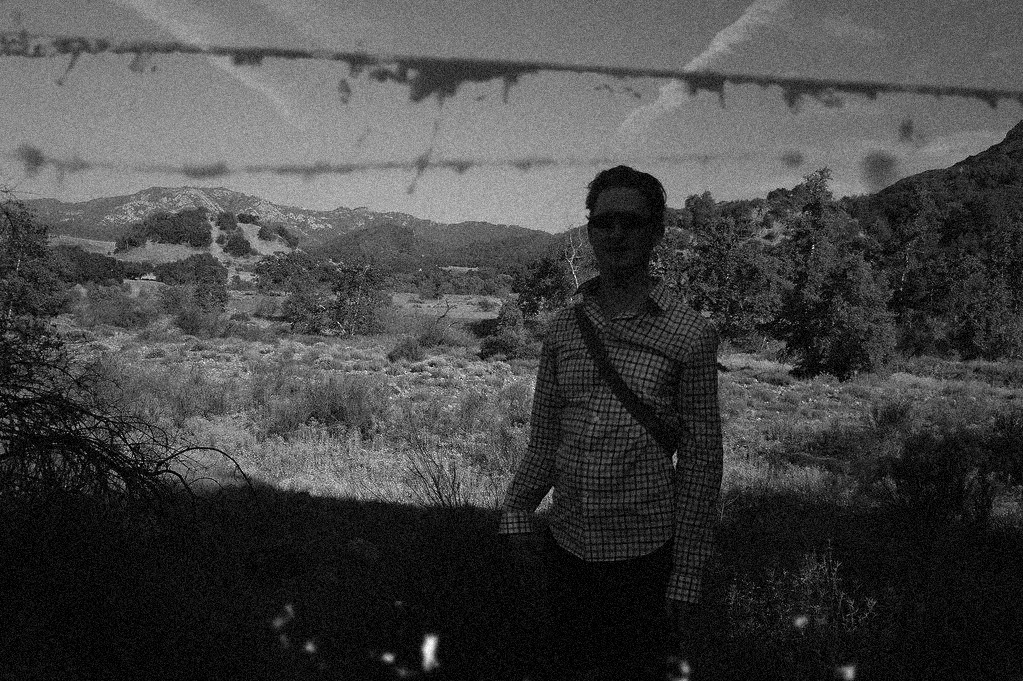} & \ddfig{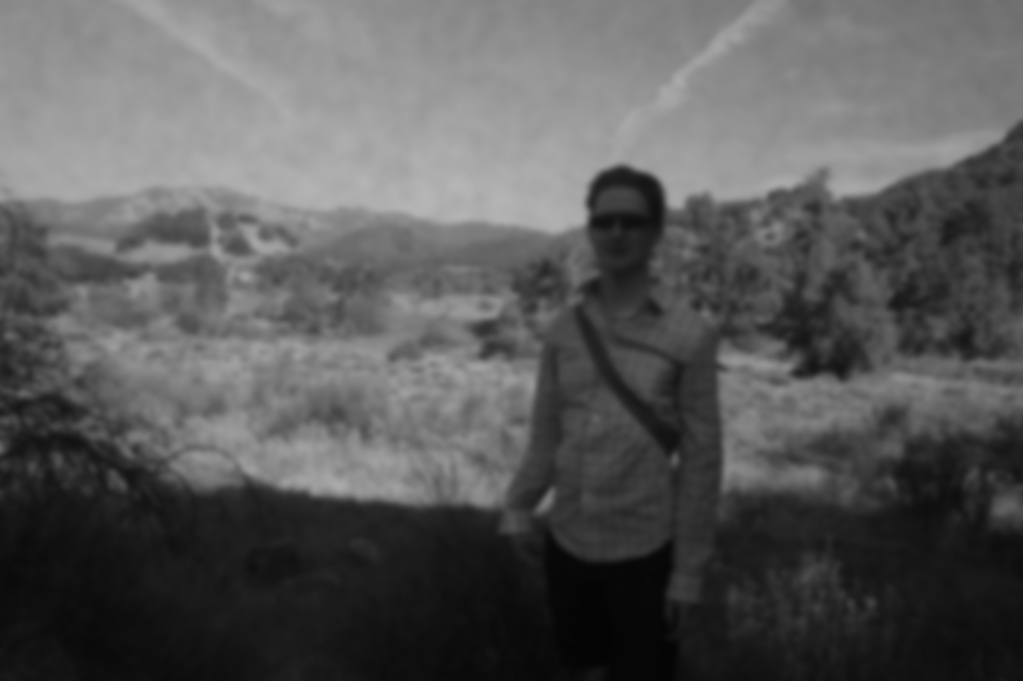} \\
      (c) Deteriorated Example 2 & (d) Deteriorated Example 3 \\
   \end{tabular}
   \setlength{\tabcolsep}{6pt}
   \vspace{-4mm}
   \caption{\rev{\textbf{Examples of synthetic deterioration effects applied to a black and
   white image.} (a) For the original image, (b-d) various types of both
   algorithm-based and example-based deterioration effects, such as JPEG
   compression artifacts and film scratches, are randomly added.} \cond{Video licensed in the Public Domain.}}
   \label{fig:deterioration}
\end{figure}

For all the noise, we use similar data augmentation techniques as used for the
input video. In particular, the noise images are scaled randomly such that the
shortest edge is between $[256,720]$ pixels, both horizontally and vertically
flipped with 50\% probability, rotated randomly between $[-5,5]$ degrees,
cropped to $256\times 256$ pixels, randomly scaled by $\mathcal{U}(0.5, 1.5)$,
and randomly either subtracted or added to the original image. \cond{Some
generated training examples are shown in Fig.~\ref{fig:deterioration}.}

\subsection{Optimization}

Training is initially done of the pre-processing network and source-reference
network separately for 500,000 iterations. Afterwards, they are trained
together in an end-to-end fashion by optimizing Eq.~\eqref{eq:optimization}.
For the optimization method, we rely on the ADADELTA
algorithm~\cite{ZeilerARXIV2012}, which is a variant of stochastic gradient
descent which heuristically estimates the learning rate parameter, thus
requiring no hyper-parameters to tune.

\section{Results}

We train our model on our dataset with $\gamma=10^{-4}$ and a batch-size of 20.
We use the model with the lowest validation loss as our final model. We
evaluate both quantitative and qualitative and compare with existing methods.


\subsection{Comparison with Existing Approaches}

We compare the results of our approach with both existing approaches and strong
baselines with a quantitative evaluation.
In particular, for restoration, we compare against the approach
of~\cite{ZhangTIP2017} \rev{and \cite{YuARXIV2018}}, and
for colorization we compare against the propagation-based approach
of~\cite{VondrickECCV2018} \rev{and single-image interactive approach of
\cite{ZhangSIGGRAPH2017}}. For both remastering, \ie, joint restoration
and colorization, we compare against \cond{all possible combinations of restoration and colorization approaches, \eg, the combination of \cite{ZhangTIP2017} and
\cite{VondrickECCV2018} used together.} The approach of~\cite{ZhangTIP2017}
\rev{and \cite{YuARXIV2018}} consists of a deep residual
convolutional neural network for single image restoration. \rev{We note that
the approach of~\cite{YuARXIV2018} is an extension of~\cite{FanCVPRW2018} and
winner of the NTIRE 2018 super resolution image
challenge\footnote{\url{http://www.vision.ee.ethz.ch/ntire18/}}. We modified
the model of \cite{YuARXIV2018} by removing the up-sampling layer at the end as
the target task is restoration and not super-resolution.}
The approach of~\cite{VondrickECCV2018} is a recursive convolutional neural
network that can propagate color information. The approach
of~\cite{ZhangSIGGRAPH2017} is a single-image convolutional neural network
approach that can use user-provided hints, \cond{which we use to provide the reference image color information.}
We also compare against \rev{two} strong colorization
baselines consisting of our proposed model with the temporal convolution layers
replaced with spatial convolution layers, and \rev{of our proposed model}
without self-attention layers. \rev{For restoration,} we compare to a baseline
consisting of our pre-processing network without the skip connection.
\rev{Finally, we also compare against a baseline consisting of the restoration
and colorization networks of our approach trained independently, \ie, without
joint training.} All approaches are trained using exactly the same training
data and training approach for fair comparison.

We compare using our test set consisting of 300 videos from the Youtube-8M
dataset. For each video we randomly sample a subset of either 90 or 300 frames,
and use the subset as the ground truth. Given that these videos are not noisy
nor degraded, we follow the same approach for generating training data to
generate deteriorated inputs for evaluation. For the example-based deterioration
effects, we use a different set of images from those of the training set to
evaluate generalization.
\rev{We use Peak Signal-to-Noise Ratio (PSNR) as an evaluation metric, and
compute the PSNR using the luminance channel only for the restoration task,
using the chrominance channels only for the colorization task, and using all
the image channels for the remastering task.}

For the reference color images, in the case of the 90 frame subset, we only provide
the first frame as a reference image, while in the case of the 300 frame
subset, we provide every 60th frame starting from the first frame as a
reference image. For our approach, all the reference frames are provided at all
times. In the case of the approach of~\cite{VondrickECCV2018}, as it only
propagates the color and is unable to naturally handle multiple reference
images, we replace the output image with the new reference image when
necessary as shown in Fig.~\ref{fig:recursive}. We note that the same random
subset of all videos is used for all the approaches.

\subsubsection{Remastering Results}

\newcommand{\yfig}[1]{
   \includegraphics[width=0.19\linewidth]{figs/comparisons/synthetic/#1_in}&
   \includegraphics[width=0.19\linewidth]{figs/comparisons/synthetic/#1_c_zz}&   
   \includegraphics[width=0.19\linewidth]{figs/comparisons/synthetic/#1_c_zv}&
   \includegraphics[width=0.19\linewidth]{figs/comparisons/synthetic/#1_c_ours}&
   \includegraphics[width=0.19\linewidth]{figs/comparisons/synthetic/#1_gt}}
\begin{figure*}[t]
   \centering
   \setlength{\tabcolsep}{1pt}
   \begin{tabular}{ccccc}
      \yfig{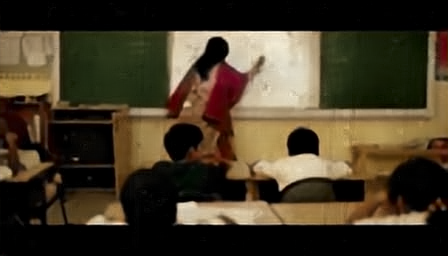} \\[-1mm]
      & \cond{24.17dB} & \cond{25.19dB} & \cond{28.12dB} & \\
      \yfig{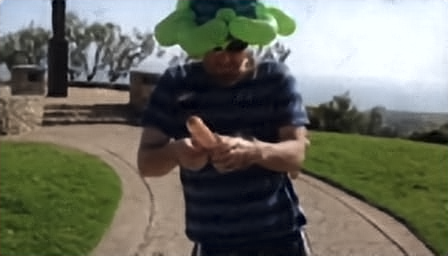} \\[-1mm]
      & \cond{27.15dB} & \cond{26.64dB} & \cond{29.63dB} & \\      
      \yfig{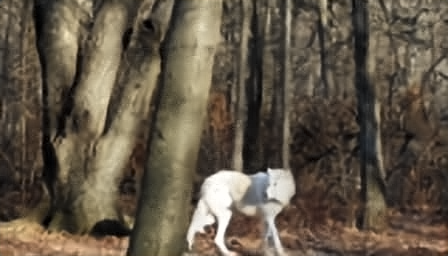} \\[-1mm]
      & \cond{24.49dB} & \cond{24.73dB} & \cond{27.71dB} & \\      
      Input & Zhang+\&Zhang+ & Zhang+\&Vondrick+ & Ours & Ground Truth \\
   \end{tabular}
   \setlength{\tabcolsep}{6pt}
   \vspace{-3mm}
   \caption{\textbf{Randomly sampled examples from the Youtube-8M test dataset with
   degradation noise.} We show one frame from several examples and compare our
   approach with the combined approach of \cite{ZhangTIP2017}
   \cond{andk \cite{ZhangSIGGRAPH2017}, and \cite{ZhangTIP2017} and
   \cite{VondrickECCV2018}}. First column shows the input frame which has been
   deteriorated with noise, the next two columns correspond to the remastering
   results with both approaches, and the last column shows the ground truth
   video. \cond{The PSNR of each approach is shown below each image. Videos
   courtesy of Naa Creation (top), Balloon Sage (middle), and Mayda Tapanes
   (bottom) and licensed under CC-by.}}
   \label{fig:synthetic_remaster}
\end{figure*}

As there is not a single approach that can handle the remastering of videos, we
compare against a pipeline approach of first processing the video with the
method of \cond{either} \cite{ZhangTIP2017} \cond{or \cite{YuARXIV2018}}, and
then propagating the reference color on the output with the approach of
\cond{either} \cite{VondrickECCV2018} \cond{or \cite{ZhangSIGGRAPH2017}}.
\rev{We also provide results of a baseline consisting of our full approach
without the joint training, \ie, the restoration and colorization networks are
trained independently.} Results are shown in Table~\ref{tbl:remastering}.
\cond{Of the pipeline-based approaches, we find that, while they have similar
performance, the combination of \cite{ZhangTIP2017} and \cite{ZhangSIGGRAPH2017}
gives the highest performance. However,} our approach outperforms the existing
pipeline based approaches \rev{and the strong baseline that doesn't use joint
training}. \rev{This shows that even though the restoration and colorization
models are first trained independently before being further trained jointly,
the joint training plays an important role in improving the quality of the
final results.} It is also interesting to point out that while the performance
of existing approaches degrades with longer videos and more reference color
images, our approach improves in performance. This is likely due to all the
reference color images being used to remaster each frame.  Several randomly
chosen examples are shown in Fig.~\ref{fig:synthetic_remaster}, where we can see that
existing approaches fail to both remove the noise and propagate the color,
while our approach performs well in both cases.

\begin{table}
   \caption{\textbf{Quantitative remastering results.} We compare the results of our
   model with that of restoring each frame with the approach
   of~\cite{ZhangTIP2017}, and propagating reference color with the approach
   of~\cite{VondrickECCV2018} on synthetically deteriorated videos from the
   Youtube-8M dataset, \rev{and with a baseline that consists of our model
   without using joint training}. We perform two types of experiments: one
   using a random 90-frame subset from each video with 1 reference frame, and
   one using a random 300-frame subset with 5 reference frames.}
   \label{tbl:remastering}
   \vspace{-2mm}
   \begin{tabular}{rccc}
      \toprule
      Approach & Frames & \# Ref. & PSNR \\
      \midrule
      \cond{Zhang+\shortcite{ZhangTIP2017}\&Zhang+\shortcite{ZhangSIGGRAPH2017}} & \cond{90} & \cond{1} & \cond{27.13} \\
                              & \cond{300} & \cond{5} & \cond{27.31} \\[2mm]
      \cond{Yu+\shortcite{YuARXIV2018}\&Zhang+\shortcite{ZhangSIGGRAPH2017}} & \cond{90} & \cond{1} & \cond{26.43} \\
                              & \cond{300} & \cond{5} & \cond{26.59} \\[2mm]
      Zhang+\shortcite{ZhangTIP2017}\&Vondrick+\shortcite{VondrickECCV2018} &  90 & 1 & 26.43 \\
                              & 300 & 5 & 26.60 \\[2mm]
      \cond{Yu+\shortcite{YuARXIV2018}\&Vondrick+\shortcite{VondrickECCV2018}} & \cond{90} & \cond{1} & \cond{26.85} \\
                              & \cond{300} & \cond{5} & \cond{26.89} \\[2mm]
      \rev{Ours \nicefrac{w}{o} joint training} &  90 & 1 & 29.07 \\
                              & 300 & 5 & 29.23 \\[2mm]   
      Ours                    &  90 & 1 & \bf{30.83} \\
                              & 300 & 5 & \bf{31.14} \\
      \bottomrule
   \end{tabular}
\end{table}

\subsubsection{Restoration Results}

\newcommand{\rfig}[1]{%
   \includegraphics[width=0.19\linewidth]{figs/comparisons/restoration/#1_in}&
   \includegraphics[width=0.19\linewidth]{figs/comparisons/restoration/#1_zhang}&
   \includegraphics[width=0.19\linewidth]{figs/comparisons/restoration/#1_wdsr}&
   \includegraphics[width=0.19\linewidth]{figs/comparisons/restoration/#1_ours}&
   \includegraphics[width=0.19\linewidth]{figs/comparisons/restoration/#1_gt}}
\begin{figure*}[t]
   \centering
   \setlength{\tabcolsep}{1pt}
   \begin{tabular}{ccccc}
      \rfig{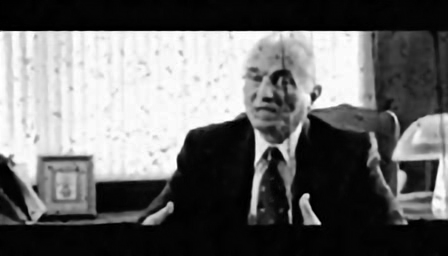} \\
      \rfig{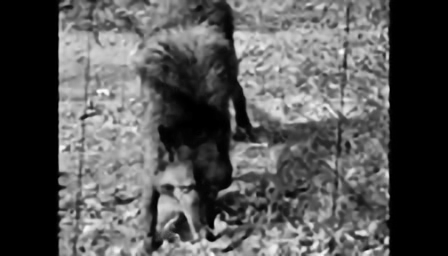} \\
      Input & \cite{ZhangTIP2017} & \cite{YuARXIV2018} & Ours & Ground Truth \\
   \end{tabular}
   \setlength{\tabcolsep}{6pt}
   \vspace{-3mm}
   \caption{\rev{\textbf{Restoration results on the Youtube-8M test dataset
   with degradation noise.} We show one frame from several examples and compare
   our approach with the approaches of~\cite{ZhangTIP2017}
   and~\cite{YuARXIV2018}.  The first column shows the input frame which has
   been deteriorated with noise, the next three columns correspond to the black
   and white restoration of each approach, and the last column corresponds to
   the ground truth video. \cond{Videos courtesy of Naa Creation (top), and Mayda
   Tapanes (bottom) and licensed under CC-by.}}}
   \label{fig:synthetic_restoration}
\end{figure*}

We compare our approach with that of~\cite{ZhangTIP2017},
\rev{\cite{YuARXIV2018}}, and a baseline for video
restoration. The baseline consists of our pre-processing model without the skip
connection that adds the input to the output. As color is not added, no
reference color images are provided and the evaluation is done using only the
300 frame subset. Results are shown in Table~\ref{tbl:restoration}. We can see
that the baseline, the approach of \cite{ZhangTIP2017}, \rev{and the approach
of \cite{YuARXIV2018}} perform similarly, while our full
pre-processing model, with a skip connection, outperforms both.
\cond{Example results are shown in Fig.~\ref{fig:synthetic_restoration}.}

\begin{table}
   \caption{\textbf{Quantitative restoration results.} We compare the results of our
   pre-processing network with the approach of~\cite{ZhangTIP2017},
   \rev{\cite{YuARXIV2018}, and a baseline of our
   approach without the skip connection} for restoring synthetically
   deteriorated videos from the Youtube-8M dataset.}
   \label{tbl:restoration}
   \vspace{-2mm}
   \begin{tabular}{rccc}
      \toprule
      Approach & Frames & \# Ref. & PSNR \\
      \midrule
      \cite{ZhangTIP2017} & 300 & - & 25.08 \\
      \rev{\cite{YuARXIV2018}} & 300 & - & 24.49 \\[2mm]
      \rev{Ours \nicefrac{w}{o} skip connection} & 300 & - & 24.73\\
      Ours                & 300 & - & \bf{26.13} \\
      \bottomrule
   \end{tabular}
\end{table}

\subsubsection{Colorization Results}

\newcommand{\cfig}[1]{%
   \includegraphics[width=0.19\linewidth]{figs/comparisons/colorization/#1_in}&
   \includegraphics[width=0.19\linewidth]{figs/comparisons/colorization/#1_zhang2}&
   \includegraphics[width=0.19\linewidth]{figs/comparisons/colorization/#1_von}&
   \includegraphics[width=0.19\linewidth]{figs/comparisons/colorization/#1_ours}&
   \includegraphics[width=0.19\linewidth]{figs/comparisons/colorization/#1_ref}}
\begin{figure*}[t]
   \centering
   \setlength{\tabcolsep}{1pt}
   \begin{tabular}{ccccc}
      \cfig{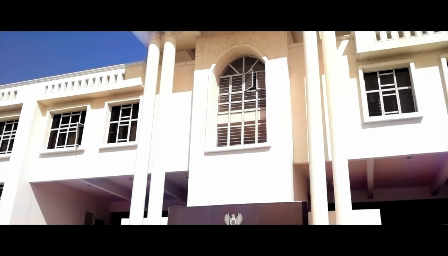} \\
      \cfig{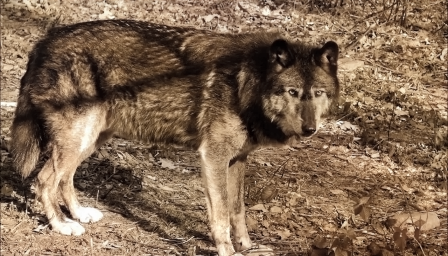} \\
      Input & \cite{ZhangSIGGRAPH2017} & \cite{VondrickECCV2018} & Ours & Reference \\
   \end{tabular}
   \setlength{\tabcolsep}{6pt}
   \vspace{-3mm}
   \caption{\rev{\textbf{Colorization results on the Youtube-8M test dataset}.
   We show one frame from several examples and compare our
   approach with the colorization approach of~\cite{ZhangSIGGRAPH2017}
   without using the reference image and the RNN-based approach~\cite{VondrickECCV2018}
   which uses the reference image.
   The first column shows the input frame, the next three columns correspond to the
   colorization of each approach, and the last corresponds to the reference image.
   Note that the input frame is not the same frame as the reference image.}
   \cond{Videos courtesy of Naa Creation (top), and Mayda Tapanes (bottom) and
   licensed under CC-by.}}
   \label{fig:synthetic_color}
\end{figure*}

We compare against the approach of \rev{\cite{ZhangSIGGRAPH2017} \cond{using
global hints}, the approach of} \cite{VondrickECCV2018} and \rev{two baselines:
one consisting of our source-reference network without temporal convolutions
and one without self-attention for colorization.} Results are shown in
Table~\ref{tbl:colorization}, and we can see that our approach outperforms
existing approaches and the baselines. Similar to the remastering case, our
approach performs significantly better on longer videos with additional
references images, which is indicative of the capabilities of the
source-reference attention: not only is it possible to colorize long sequences
with many reference images, it is beneficial for performance. \rev{An
interesting result is that self-attention plays a critical role in our model.
We believe this is due to the fact it allows each output pixel to be computed
using information from the entire image, which would require many more
convolutional layers if self-attention was not employed.} \cond{Example results
are shown in Fig.~\ref{fig:synthetic_color}.}

\begin{table}
   \caption{\textbf{Quantitative colorization results.} We compare the
   colorization results of our source-reference network with the approach of
   \cond{\cite{ZhangSIGGRAPH2017} using global hints, and}
   \cite{VondrickECCV2018} for the colorization of videos from the Youtube-8M
   dataset. We perform two types of experiments: one using a random 90-frame
   subset from each video with 1 reference frame, and one using a random
   300-frame subset with 5 reference frames.}
   \label{tbl:colorization}
   \vspace{-2mm}
   \begin{tabular}{rccc}
      \toprule
      Approach & Frames & \# Ref. & PSNR \\
      \midrule
      \cond{\cite{ZhangSIGGRAPH2017}}& \cond{90} & \cond{1} & \cond{31.28} \\
                              & \cond{300} & \cond{5} & \cond{31.16} \\[2mm]
      \cite{VondrickECCV2018} &  90 & 1 & 31.55 \\
                              & 300 & 5 & 31.70 \\[2mm]
      \rev{Ours \nicefrac{w}{o} temporal conv.} &  90 & 1 & 28.46 \\
                              & 300 & 5 & 28.51 \\[2mm]
      \rev{Ours \nicefrac{w}{o} self-attention} &  90 & 1 & 29.00 \\
                              & 300 & 5 & 28.72 \\[2mm]                              
      Ours                    &  90 & 1 & \bf{34.94} \\
                              & 300 & 5 & \bf{36.26} \\
      \bottomrule
   \end{tabular}
\end{table}

\subsection{Qualitative Results}

\newcommand{\side}[1]{\rotatebox{90}{#1}\hspace{2pt}}
\newcommand{\qfig}[1]{\includegraphics[width=0.31\linewidth]{figs/comparisons/#1}}
\newcommand{\reffig}[2]{\includegraphics[width=#2\linewidth]{figs/comparisons/#1}}
\begin{figure*}
   \begin{subfigure}{0.495\linewidth}
   \centering
   \setlength{\tabcolsep}{1pt}
   \begin{tabular}{rccc}
      \side{References} &
         \qfig{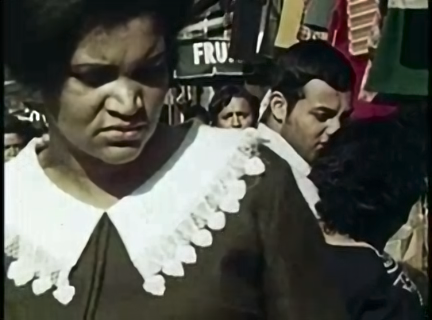} &
         \qfig{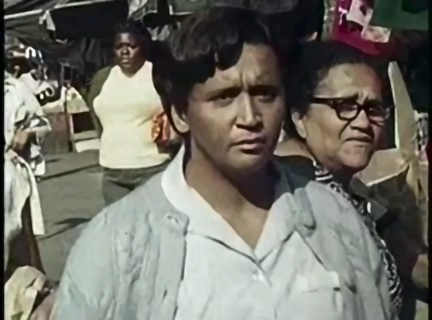} &
         \qfig{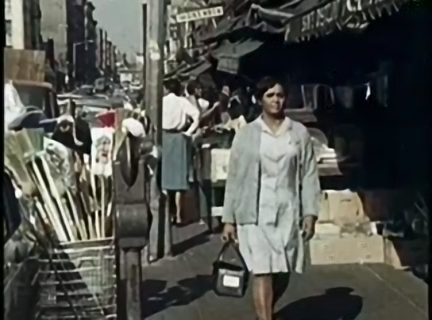} \\[-1mm]
      & $t=0$ & $t=150$ & $t=272$ \\[1mm]
      \side{$t=5$} &
         \qfig{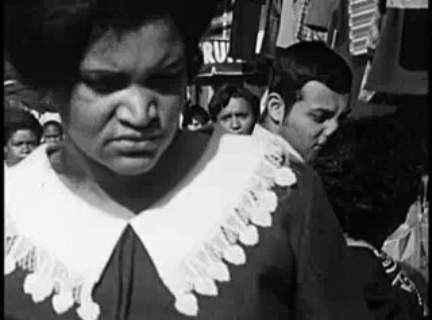} &
         \qfig{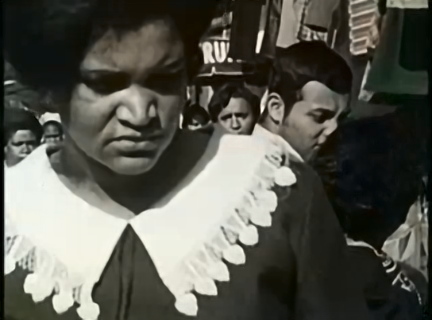} &
         \qfig{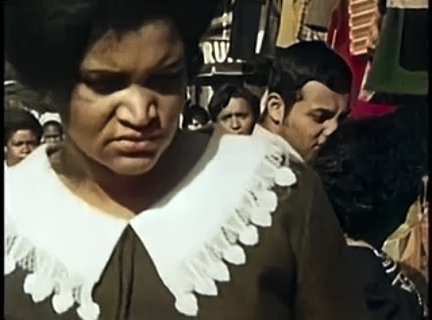} \\
      \side{$t=85$} &
         \qfig{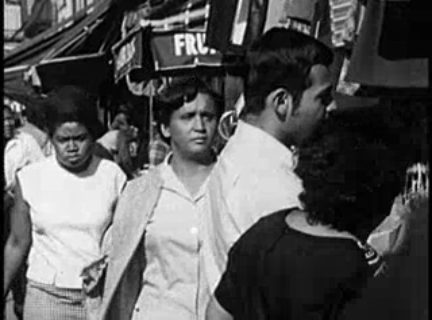} &
         \qfig{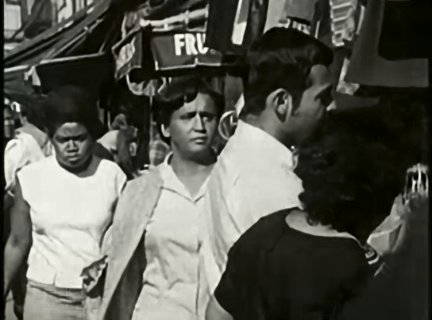} &
         \qfig{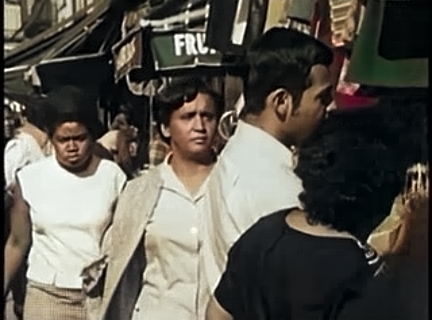} \\
      \side{$t=302$} &
         \qfig{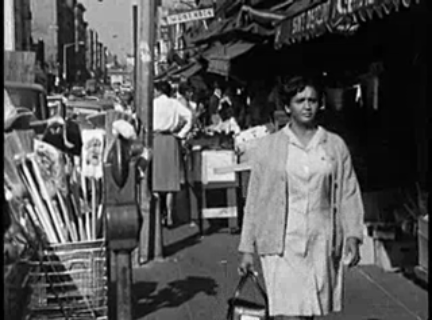} &
         \qfig{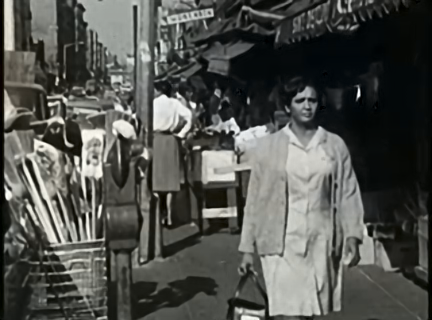} &
         \qfig{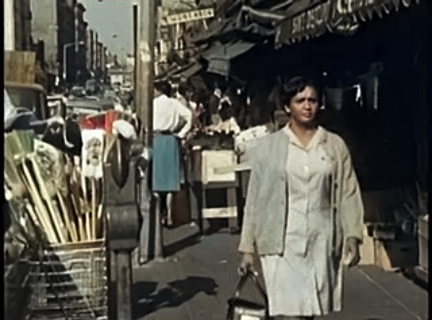} \\
      & Input & Zhang+\&Vondrick+ & Ours \\
   \end{tabular}
   \setlength{\tabcolsep}{6pt}
   \vspace{-2mm}
   \caption{``Right to Health, A (Part I)'' (Unknown).}
   \label{fig:right_to_health}
   \end{subfigure}
   \vspace{2mm}
   \begin{subfigure}{0.495\linewidth}
   \centering
   \setlength{\tabcolsep}{1pt}
   \begin{tabular}{rccc}
      \side{References} &
         \qfig{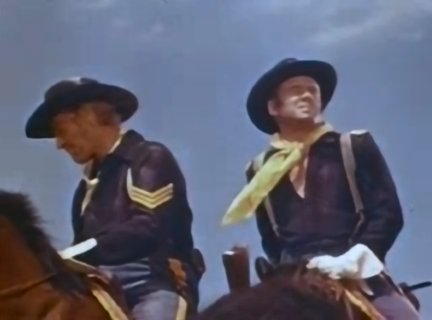} &
         \qfig{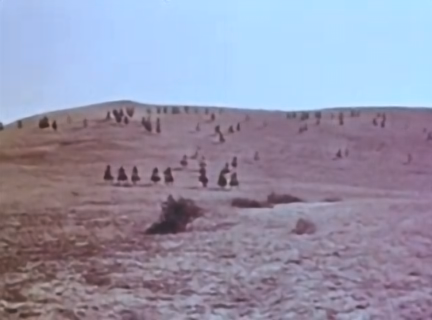} \\[-1mm]
      & $t=0$ & $t=888$ \\[1mm]
      \side{$t=5$} &
         \qfig{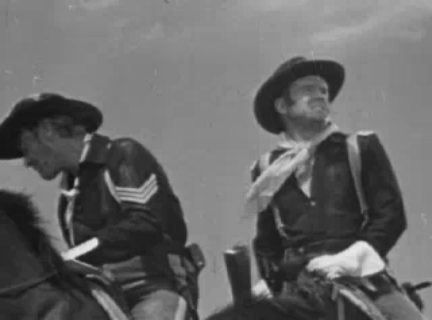} &
         \qfig{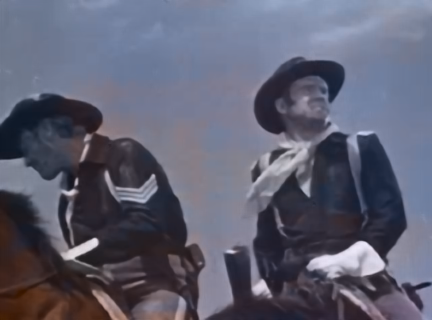} &
         \qfig{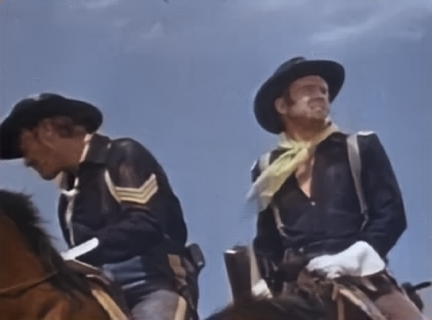} \\
      \side{$t=800$} &
         \qfig{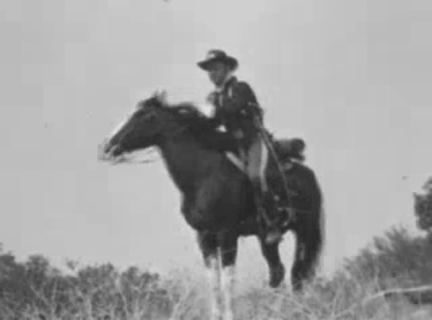} &
         \qfig{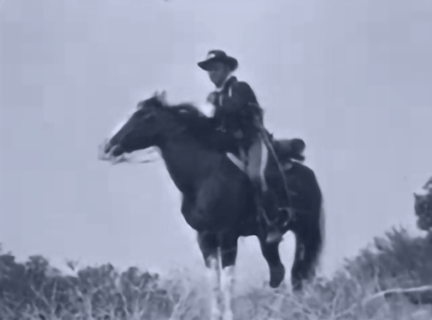} &
         \qfig{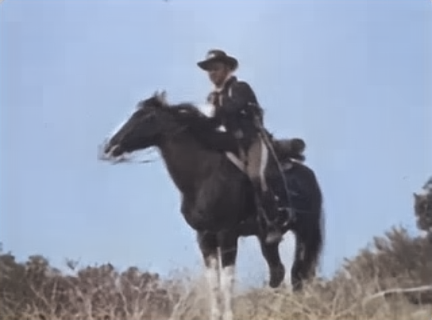} \\
      \side{$t=893$} &
         \qfig{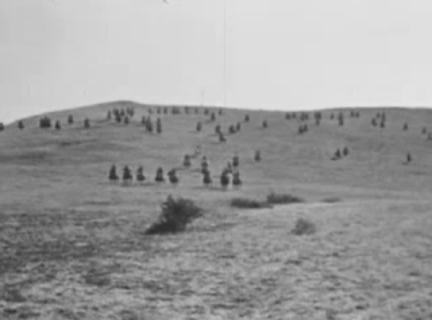} &
         \qfig{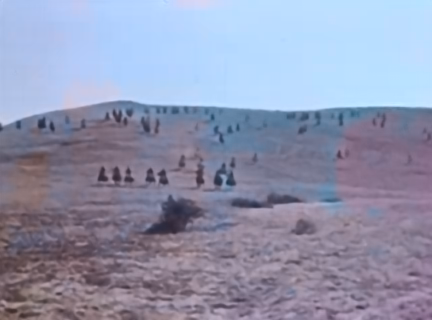} &
         \qfig{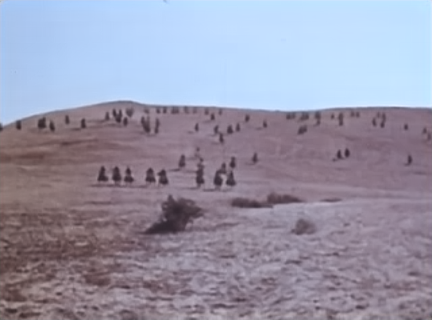} \\
      & Input & Zhang+\&Vondrick+ & Ours \\
   \end{tabular}
   \setlength{\tabcolsep}{6pt}
   \vspace{-2mm}
   \caption{``Freedom Highway'' (1956).}
   \end{subfigure}
   \vspace{2mm}
   \begin{subfigure}{0.495\linewidth}
   \centering
   \setlength{\tabcolsep}{1pt}
   \begin{tabular}{rccccc}
      \side{References} &
         \reffig{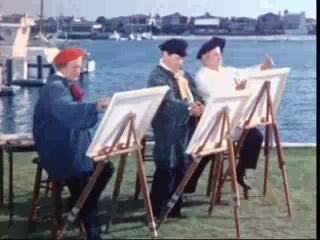}{0.18} &
         \reffig{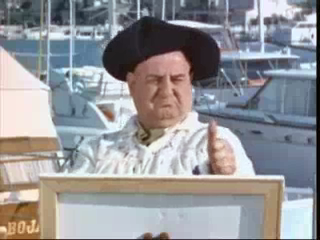}{0.18} &
         \reffig{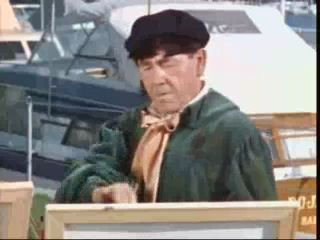}{0.18} &
         \reffig{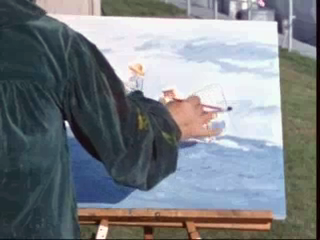}{0.18} &
         \reffig{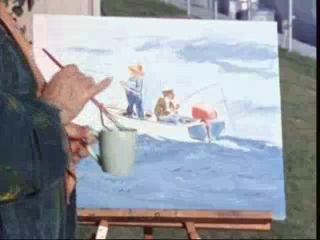}{0.18} \\[-1mm]
      & $t=12$ & $t=180$ & $t=196$ & $t=312$ & $t=348$ \\[1mm]
   \end{tabular}
   \begin{tabular}{rccc}
      \side{$t=14$} &
         \qfig{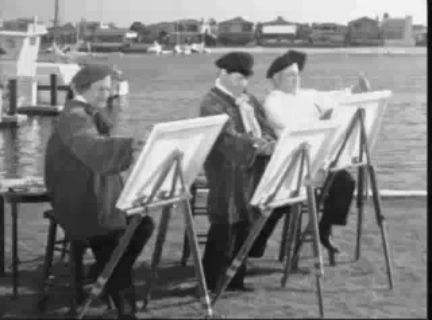} &
         \qfig{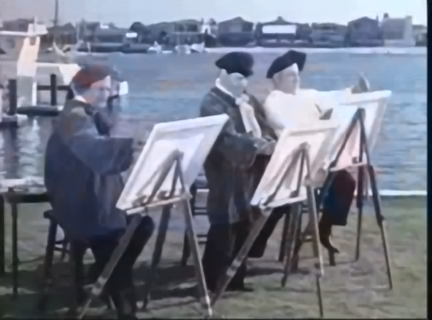} &
         \qfig{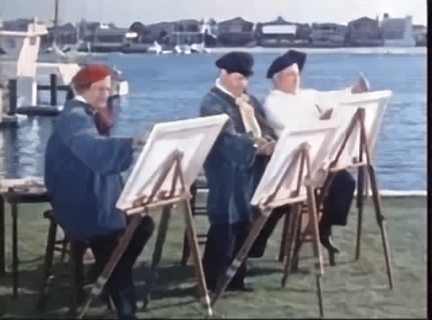} \\
      \side{$t=263$} &
         \qfig{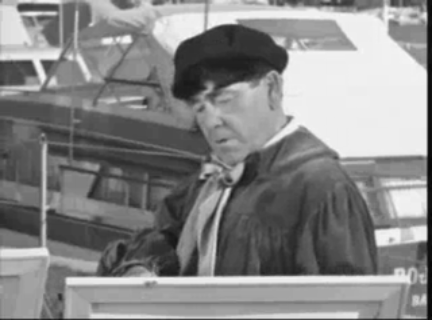} &
         \qfig{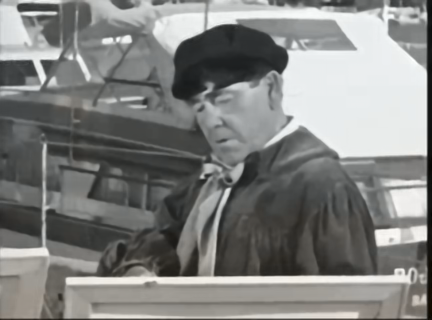} &
         \qfig{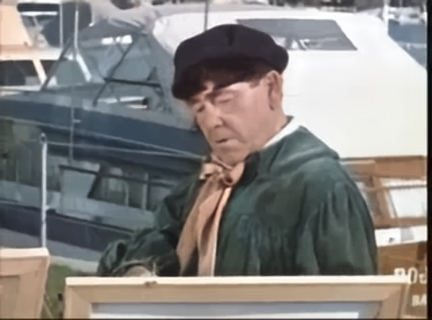} \\
      \side{$t=320$} &
         \qfig{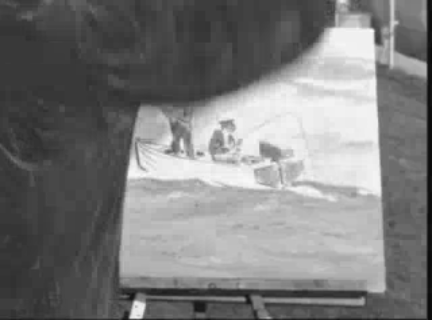} &
         \qfig{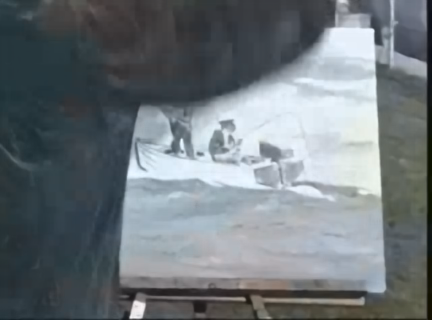} &
         \qfig{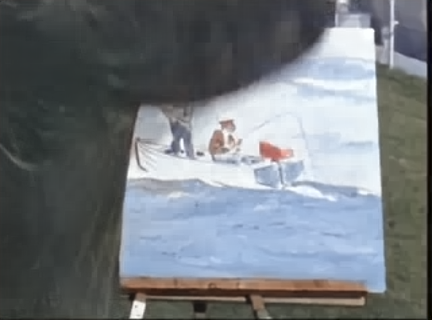} \\
      & \cond{Input} & \cond{Zhang+\&Vondrick+} & \cond{Ours} \\
   \end{tabular}
   \setlength{\tabcolsep}{6pt}
   \vspace{-2mm}
   \caption{\cond{``Color Craziness'' (1965).}}
   \end{subfigure}
   \begin{subfigure}{0.495\linewidth}
   \centering
   \setlength{\tabcolsep}{1pt}
   \begin{tabular}{rccccccc}
      \multirow{2}{*}{\side{References}} &
         \reffig{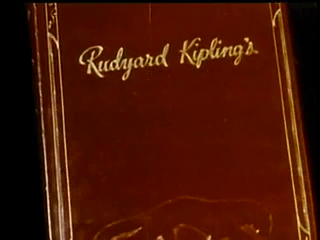}{0.15} &
         \reffig{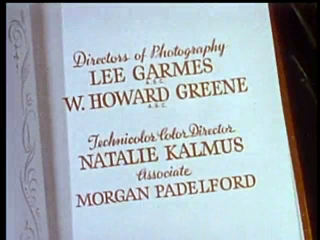}{0.15} &
         \reffig{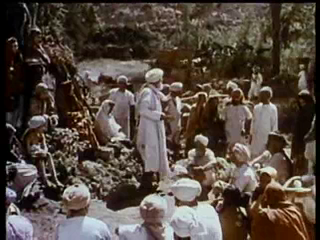}{0.15} &
         \reffig{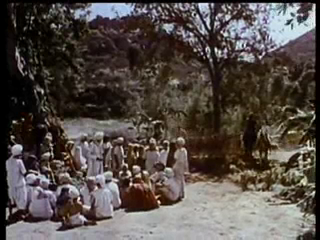}{0.15} &
         \reffig{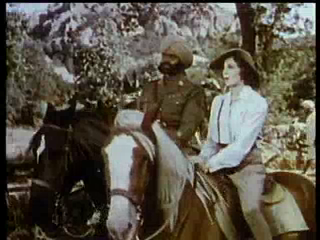}{0.15} &
         \reffig{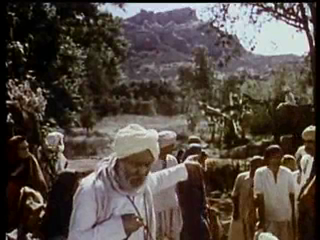}{0.15} \\[-1mm]
      & $t=155$ & $t=1043$ & $t=2582$ & $t=2831$ & $t=2973$ & $t=3539$ \\[1mm]
      & \raisebox{3mm}{$\cdots$} &
         \reffig{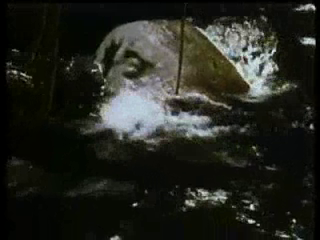}{0.15} &
         \reffig{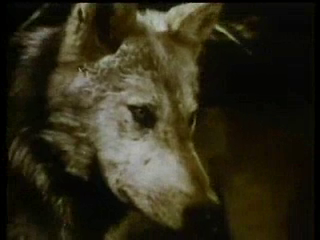}{0.15} &
         \reffig{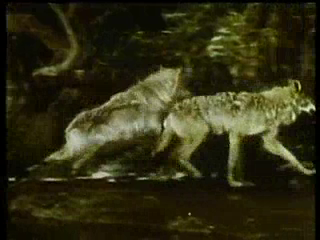}{0.15} &
         \reffig{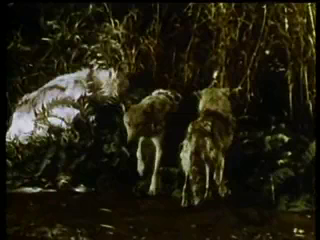}{0.15} &         
         \reffig{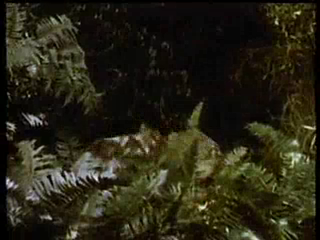}{0.15} \\[-1mm]
      &   & $t=8878$ & $t=9043$ & $t=9197$ & $t=9337$ & $t=9659$ \\[1mm]
   \end{tabular}
   \begin{tabular}{rccc}
      \side{$t=453$} &
         \qfig{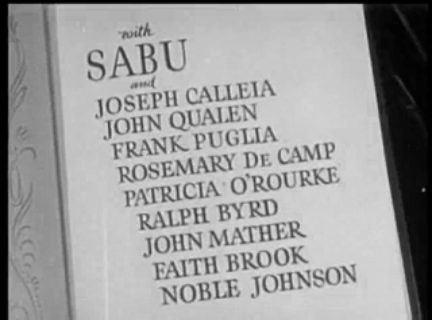} &
         \qfig{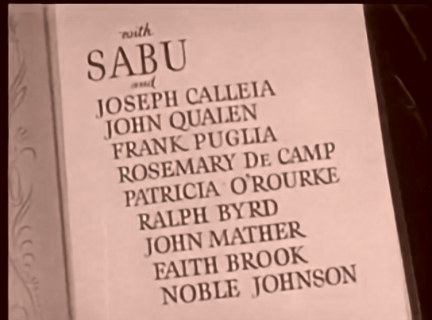} &
         \qfig{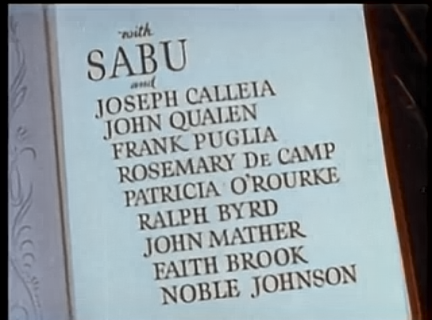} \\
      \side{$t=4771$} &
         \qfig{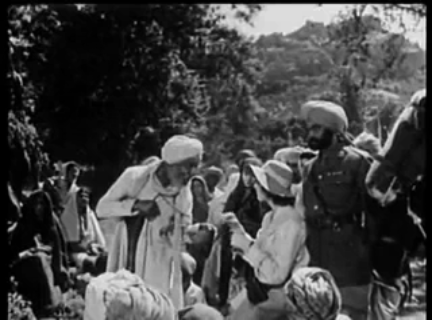} &
         \qfig{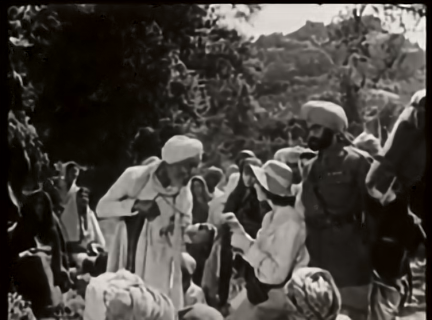} &
         \qfig{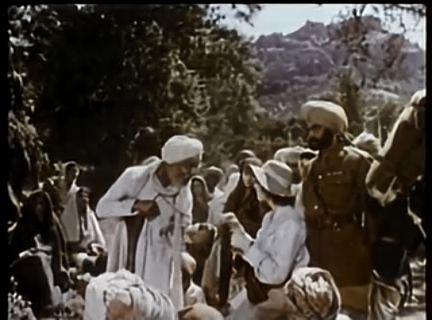} \\
      \side{$t=9023$} &
         \qfig{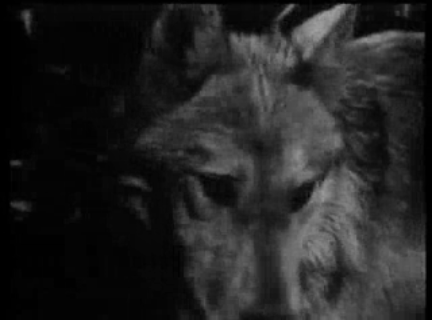} &
         \qfig{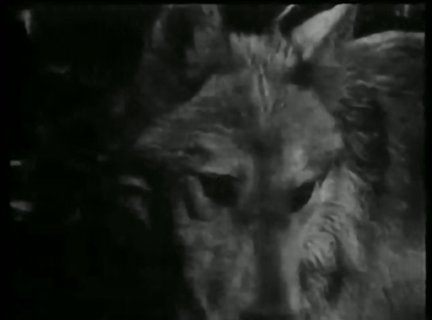} &
         \qfig{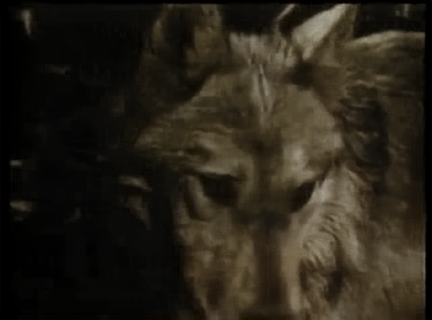} \\
      & \cond{Input} & \cond{Zhang+\&Vondrick+} & \cond{Ours} \\
   \end{tabular}
   \setlength{\tabcolsep}{6pt}
   \vspace{-2mm}
   \caption{\cond{``The Jungle Book'' (1967).}}
   \end{subfigure}     
   \vspace{-3mm}
   \caption{\textbf{Qualitative comparison with the combined approach of
   Zhang+\shortcite{ZhangTIP2017} and Vondrick+\shortcite{VondrickECCV2018}.}
   We show the reference color images in the first row with their timestamps.
   Afterwards four different frames taken from the input video and output
   videos are shown. \cond{Note that the example of (d) is remastered with 41 reference images of which we only show a subset.}
   ``Right to Health, A (Part I)'', ``Freedom
   Highway'', \cond{``Color Craziness'', and ``The Jungle Book''} are licensed in the public domain.}
   \label{fig:qualitative}
\end{figure*}

We show qualitative results in Fig.~\ref{fig:qualitative} on diverse challenging
real world vintage film examples. As the videos are originally color, we use
images from the original video as the reference images, and then compare both our
remastering approach and a pipeline approach of denoising with the approach
of~\cite{ZhangTIP2017} and then adding color with the method
of~\cite{VondrickECCV2018}. We can see how our approach is able
to perform a consistent remastering, while existing approaches lose track of
the colorization and fail to produce pleasing results, which is consistent with
our quantitative evaluation.

We also perform a qualitative comparison of restoration results on vintage film
in Fig.~\ref{fig:restorationqual} with the approach of~\cite{ZhangTIP2017}. We
can see how the approach of~\cite{ZhangTIP2017} can restore small noise, but
fails at larger noise. Our approach is able to handle both small and large
noise, while also sharpening the input image.

\cond{
\subsection{Computation Time}

For a $528\times 400$-px input video, our approach takes 69ms per frame with a Nvidia GTX 1080Ti GPU, with 4ms corresponding to the restoration stage, and 65ms corresponding to the colorization stage.
}

\newcommand{\nfig}[1]{\includegraphics[width=0.345\linewidth]{figs/restoration/#1}}
\newcommand{\nnfig}[1]{\includegraphics[width=\linewidth]{figs/restoration/#1}}
\renewcommand{\nfig}[1]{\includegraphics[width=0.325\linewidth]{figs/restoration/#1}}
\begin{figure}
   \centering
   \setlength{\tabcolsep}{1pt}
   \begin{tabular}{ccc}
      \nfig{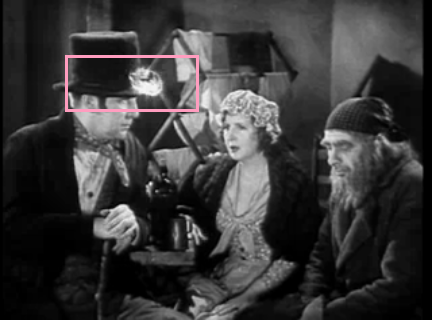} &
         \nfig{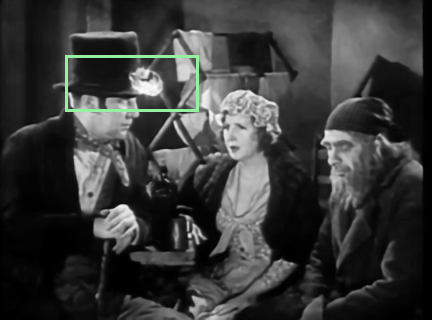} &
         \nfig{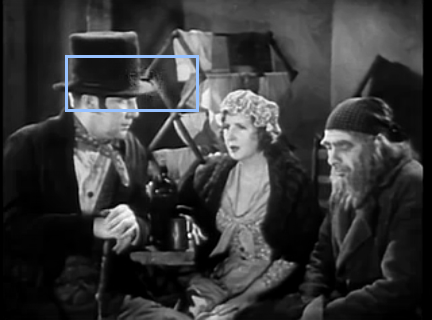} \\
      \nfig{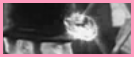} &
         \nfig{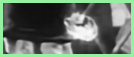} &
         \nfig{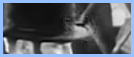}\\
      Input & \cite{ZhangTIP2017} & Ours\\
   \end{tabular}
   \setlength{\tabcolsep}{6pt}
   \vspace{-3mm}
   \caption{\rev{\textbf{Restoration result on vintage film.} We compare with
   the approach of \cite{ZhangTIP2017}, and show the boxed area zoomed in on
   the bottom row.  We can see that the relatively large noise is ``inpainted''
   with our network.} First two rows are frames taken from the movie ``Oliver
   Twist'' (1933) which is licensed in the public domain.}
   \label{fig:restorationqual}
\end{figure}

\section{Limitations and Discussion}

We have presented an approach for the remastering of vintage film based on
temporal convolutional networks with source-reference attention mechanisms that
allow for using an arbitrary number of reference color images. Although the
source-reference attention mechanism is a powerful tool to incorporate
reference images into a processing framework and is amenable to process videos
of any resolution, it suffers from $\mathcal{O}(N_rH_rW_rT_sH_sW_s)$ memory
usage. Available system memory will thus limit the maximum resolution that can
be processed. However,in practice,
as most vintage movies are stored at low resolutions due to limits of the film
technology, they do not have to be processed at resolutions that would not be
possible with attention-based mechanisms.

\newcommand{\lfig}[1]{\includegraphics[width=0.33\linewidth]{figs/limitation/#1}}
\begin{figure}
   \setlength{\tabcolsep}{1pt}
   \begin{tabular}{ccc}
      \lfig{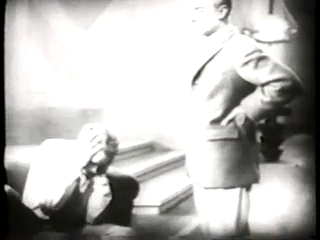} &
      \lfig{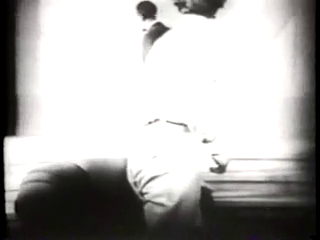} &
      \lfig{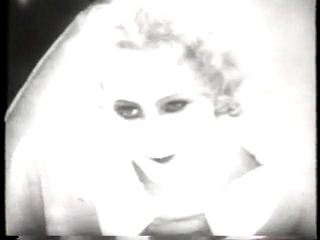} \\
      \lfig{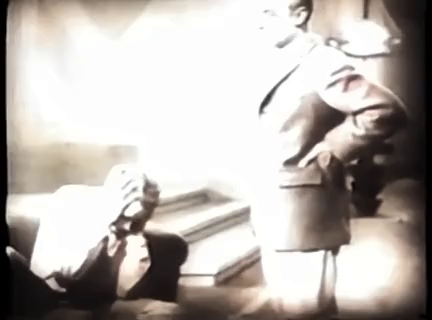} &
      \lfig{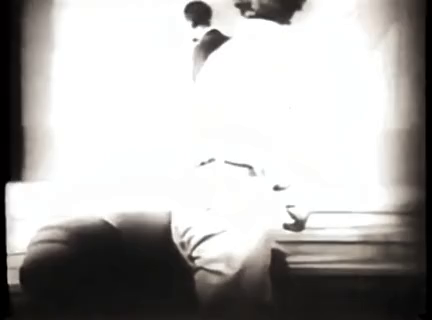} &
      \lfig{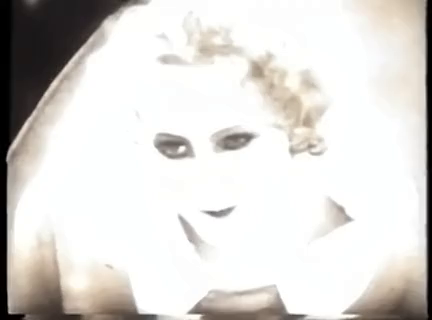} \\
   \end{tabular}
   \setlength{\tabcolsep}{6pt}
   \vspace{-4mm}
   \caption{\textbf{Limitation of our approach.} Example of severely
   deteriorated film which is not possible to remaster with the current
   approach. \cond{The first row shows frames from the original input video and
   the second row shows the output of our approach.} Images taken from the
   movie ``Metropolis'' (1925) which is licensed in the public domain.}
   \label{fig:limitation}
\end{figure}

Currently, the proposed approach relies on fully supervised learning and can
not fill missing frames nor extreme degradation that leaves a large region of
the image missing during many frames as shown in Fig.~\ref{fig:limitation}. In 
these cases there is too much missing information which makes it impossible to
remaster, they would require image completion-based approaches to remake
new plausible parts of the video, which is out of the scope of this work.

\cond{Our model has a temporal resolution of 15 frames, corresponding to
roughly half a second in most videos, which can lead to small temporal
consistencies in the output video. For reference, existing approaches use a
smaller amount such as 4 frames~\cite{VondrickECCV2018} or 10
frames~\cite{LaiECCV2018}. While it should be possible to increase the temporal
resolution, this leads to slower convergence and slower computation. While
blind video temporal consistency techniques can alleviate this
issue~\cite{BonneelSIGGRAPHASIA2015,LaiECCV2018}, we found that while they are
able to slightly improve the temporal consistency, it comes at the cost of
significantly worse results. We believe that integrating such an approach with our model and training end-to-end is a possible way to improve the temporal consistency without sacrificing the quality of the results.}

\rev{We note that despite the progress in this work on remastering vintage
film, due to the complexity of the task, it still is an open problem in
computer graphics. Unlike most of the image and video research up until now,
vintage film poses a much more difficult and realistic problem as highlighted
in Fig.~\ref{fig:oldfilm}, and we hope that this work can further stimulate
research in this topic.}




\begin{acks}
This work was partially supported by JST ACT-I (Iizuka, Grant
Number: JPMJPR16U3),  JST PRESTO (Simo-Serra, Grant Number: JPMJPR1756), and
JST CREST (Iizuka and Simo-Serra, Grant Number: JPMJCR14D1).
\end{acks}

\bibliographystyle{ACM-Reference-Format}
\bibliography{top}

\end{document}